\documentclass[utf8]{FrontiersinHarvard} % for articles in journals using the Harvard Referencing Style (Author-Date), for Frontiers Reference Styles by Journal: https://zendesk.frontiersin.org/hc/en-us/articles/360017860337-Frontiers-Reference-Styles-by-Journal
\usepackage{url,hyperref,lineno,microtype,subcaption}
\usepackage[onehalfspacing]{setspace}

\usepackage{placeins} 
\usepackage[utf8]{inputenc}
\DeclareUnicodeCharacter{200B}{ }

\usepackage{graphicx} % Add this in the preamble

\def\keyFont{\fontsize{8}{11}\helveticabold }
\def\firstAuthorLast{Sample {et~al.}} %use et al only if is more than 1 author
\def\Authors{Ang He\,$^{1,2}$, Ximei Wu\,$^{1,2}$, Xing Xu\,$^{3,4}$, Jing Chen$^{1,2}$, Xiaobin Guo$^{1,2,*}$ and Sheng Xu\,$^{1,2,*}$}
\def\Address{$^{1}$Guangdong University of Technology, Guangzhou Higher Education Mega Centre, Guangzhou, 510006, PR China \\
$^{2}$Guangdong Provincial Key Laboratory of Sensing Physics and System Integration Applications \\
$^{3}$College of Engineering, South China Agricultural University, Guangzhou, China \\
$^{4}$National Banana Industry Technology System Orchard Production Mechanization Research Laboratory
  }
\def\corrAuthor{Corresponding Author}

\def\corrEmail{guoxb@gdut.edu.cn (X.-b.G.) xusheng@gdut.edu.cn(S.-X.)}

\begin{document}
\onecolumn
\firstpage{1}

\title[Running Title]{Iterative Optimization Annotation Pipeline and ALSS-YOLO-Seg for Efficient Banana Plantation Segmentation in UAV Imagery} 

\author[\firstAuthorLast ]{\Authors} %This field will be automatically populated
\address{} %This field will be automatically populated
\correspondance{} %This field will be automatically populated

\extraAuth{}% If there are more than 1 corresponding author, comment this line and uncomment the next one.
%\extraAuth{corresponding Author2 \\ Laboratory X2, Institute X2, Department X2, Organization X2, Street X2, City X2 , State XX2 (only USA, Canada and Australia), Zip Code2, X2 Country X2, email2@uni2.edu}

\maketitle

\begin{abstract}

%%% Leave the Abstract empty if your article does not require one, please see the Summary Table for full details.
\section{}
Precise segmentation of Unmanned Aerial Vehicle (UAV)-captured images plays a vital role in tasks such as crop yield estimation and plant health assessment in banana plantations. By identifying and classifying planted areas, crop area can be calculated, which is indispensable for accurate yield predictions. However, segmenting banana plantation scenes requires a substantial amount of annotated data, and manual labeling of these images is both time-consuming and labor-intensive, limiting the development of large-scale datasets. Furthermore, challenges such as changing target sizes, complex ground backgrounds, limited computational resources, and correct identification of crop categories make segmentation even more difficult. To address these issues, we proposed a comprehensive solution. Firstly, we designed an iterative optimization annotation pipeline leveraging SAM2's zero-shot capabilities to generate high-quality segmentation annotations, thereby reducing the cost and time associated with data annotation significantly. Secondly, we developed ALSS-YOLO-Seg, an efficient lightweight segmentation model optimized for UAV imagery. The model's backbone includes an Adaptive Lightweight Channel Splitting and Shuffling (ALSS) module to improve information exchange between channels and optimize feature extraction, aiding accurate crop identification. Additionally, a Multi-Scale Channel Attention (MSCA) module combines multi-scale feature extraction with channel attention to tackle challenges of varying target sizes and complex ground backgrounds. We evaluated the zero-shot segmentation performance of SAM2 on the ADE20K and Javeri datasets. Our iterative optimization annotation pipeline demonstrated a significant reduction in manual annotation effort while achieving high-quality segmentation labeling. Extensive experiments on our custom Banana Plantation segmentation dataset show that ALSS-YOLO-Seg achieves state-of-the-art performance. Our code is openly available at https://github.com/helloworlder8/computer\_vision.

\tiny
 \keyFont{ \section{Keywords:} UAV, banana plantations, changing target sizes, complex ground backgrounds, SAM2, ALSS, MSCA} %All article types: you may provide up to 8 keywords; at least 5 are mandatory.
\end{abstract}

\section{Introduction}

In recent years, the global demand for bananas (Musa spp.) has surged, establishing bananas as one of the most critical economic crops in tropical and subtropical regions \citep{de2009emergy}. As a leading banana producer, China, particularly its southern provinces, has demonstrated significant advantages in both planting area and yield, thus playing a vital role in advancing the agricultural economy \citep{yang2022land}. For banana production, accurately monitoring plant health and predicting yields are essential for enhancing production efficiency and promoting sustainable agricultural practices \citep{de2019artificial}. However, traditional manual monitoring methods are often costly, inefficient, and lack precision, making it challenging to satisfy the demands of modern agriculture for efficient and accurate monitoring \citep{diaz2011novel}.

To address these challenges, Unmanned Aerial Vehicles (UAVs) have increasingly emerged as indispensable tools for monitoring banana plantations, due to their notable efficiency, flexibility, and cost-effectiveness \citep{zhang2022banana}. By integrating UAV technology with advanced deep learning algorithms, instance segmentation can be applied to automate the identification and classification of various regions within banana plantations. This methodology significantly improves both the precision and efficiency of monitoring large-scale agricultural environments \citep{mo2021deep}. The high-resolution imagery collected by UAVs enables rapid and comprehensive data gathering on the plantation environment, supporting critical applications such as crop yield estimation and plant health assessment. Accurate segmentation of UAV-captured images is crucial for determining crop area, which is indispensable for reliable yield predictions. Additionally, the ability to extract crop features from complex environments opens up new opportunities for advancing smart agriculture and precision farming.

However, despite the benefits of UAV-based monitoring, the segmentation of banana plantation scenes requires a substantial amount of annotated data to train effective models. Manual labeling of these images is a time-consuming and labor-intensive process, presenting a significant barrier to the development of large-scale annotated datasets. For instance, Kirillov et al. \citep{kirillov2023segment} reported that annotating a single image takes an average of 34 seconds, which is highly inefficient for large-scale agricultural monitoring. To overcome these limitations, the field of artificial intelligence has witnessed a paradigm shift with the advent of foundational models that leverage pre-training on extensive datasets. Foundational models such as ChatGPT have demonstrated superior performance in natural language processing and multimodal tasks \citep{lund2023chatting}\citep{qin2023chatgpt}, while models like CLIP \citep{radford2021learning}, ALIGN \citep{jia2021scaling}, and DALLE \citep{ramesh2021zero} have shown impressive generalization across multiple domains. Although these models are not specifically designed for image segmentation, they underscore the potential of large-scale pre-trained models in various applications.

In this context, The Segment Anything Model (SAM) \citep{kirillov2023segment} has garnered significant attention within the computer vision community. SAM is built on the Vision Transformer (VIT) architecture \citep{alexey2020image} and pre-trained on the massive SA-1B dataset, containing over 11 million images and 1 billion masks. SAM stands out for its ability to generate effective segmentation results through prompt-based interactions, showcasing robust zero-shot generalization across diverse tasks. The introduction of SAM represents a promising step toward alleviating the challenges associated with manual annotation in UAV-based agricultural monitoring, offering a more scalable approach to segmentation.

Building on this foundation, the recent advancement of SAM2 \citep{ravi2024sam}, an upgraded version of SAM, further enhances segmentation capabilities across a wide range of scenes and object types. Trained on the expansive SA-V dataset and employing a Transformer-based model, SAM2 can handle more diverse and complex segmentation tasks, demonstrating superior generalization ability. Notably, SAM2 is capable of segmenting any object in any image without the need for additional training, providing new prospects for intelligent image analysis and interpretation in challenging environments such as banana plantations. This advancement with SAM2 offers new prospects for intelligent image analysis and understanding.

Despite the impressive segmentation capabilities of SAM2, its interactive prompt-based framework requires user-provided initial prompts and manual intervention for segmentation optimization. For large-scale automatic monitoring tasks in complex environments such as banana plantations, this characteristic limits SAM2's applicability. Specifically, in images with dense plantings, complex backgrounds, and indistinct boundaries, SAM2's segmentation results heavily rely on prompts, making it less effective in fully automated segmentation tasks. Moreover, the large parameter size of SAM2 further restricts its deployment in resource-constrained environments. For instance, SAM2-b contains 80.8 million parameters, while SAM2-t has 38.9 million parameters, making them impractical for direct implementation on UAV platforms. As noted in \citep{zhao2023fast}, selecting a more lightweight model can offer a better balance between efficiency and accuracy for specific tasks. Therefore, while SAM2 excels in segmentation accuracy and generalization, its full automation application in banana plantation monitoring faces significant challenges, necessitating the development of novel methods to overcome these limitations.

This research presents an iterative optimization annotation pipeline for generating segmentation masks. Initially, a few sample prompts are used with SAM2 to create initial masks and compute the corresponding Minimum Bounding Boxes (MBB). These MBB are then employed to train a detection model, which generates new detection boundaries for novel images and provides feedback to SAM2 to produce refined masks. Through iterative optimization process, our approach significantly reduces the manual annotation workload while generating high-quality segmentation masks for banana plantations.

Once high-quality segmentation masks are obtained, the next step is to train a lightweight and efficient instance segmentation model specifically designed for the banana plantation scene to meet the needs of UAV platform deployment, which can be regarded as a fully coupled knowledge distillation process of SAM2 \citep{zhang2023faster}.

In the realm of instance segmentation, current state-of-the-art approaches can be broadly classified into two categories: two-stage and single-stage algorithms. Two-stage models, such as Mask R-CNN \citep{he2017mask}, Cascade R-CNN \citep{cai2019cascade}, and Mask Scoring R-CNN \citep{huang2019mask}, first generate region proposals and then perform segmentation, typically offering higher accuracy but at the cost of increased computational complexity. In contrast, single-stage algorithms, including YOLACT \citep{bolya2019yolact}, BlendMask \citep{chen2020blendmask}, and SOLO \citep{wang2020solo}, combine detection and segmentation in a unified process, offering a more efficient alternative suitable for real-time applications.

Among the single-stage methods, the YOLO (You Only Look Once) framework has gained prominence due to its outstanding balance of speed, accuracy, and computational efficiency. YOLO's design, which enables simultaneous object detection and segmentation in a single forward pass, makes it highly suitable for tasks that demand both real-time performance and robust accuracy. Consequently, the YOLO family of algorithms has been widely applied across a diverse range of fields, including robotics, autonomous driving, and video surveillance \citep{li2023yolo, li2022cross, nguyen2021yolo}. Researchers worldwide have optimized the YOLO model to address complex agricultural production challenges.

For instance, Thakuria et al. \citep{thakuria2023improving} enhanced the YOLO model's architecture for real-time automated grading of canola health, demonstrating improvements in grading accuracy. However, challenges such as changing target sizes and the complex ground backgrounds in high-density canola fields posed difficulties, as the model sometimes struggled with occlusion, which hindered its ability to differentiate overlapping plants accurately.
Li et al. \citep{li2023tomato} developed an MHSA-YOLOv8 model for tomato maturity grading and counting, achieving commendable results in terms of accuracy. Nevertheless, real-world complexities such as variable lighting and shadows affected the model’s performance, highlighting the difficulties that arise when dealing with complex backgrounds. Additionally, variations in tomato sizes introduced further challenges, requiring robust adaptability in the segmentation process.
Chen et al. \citep{chen2024yolov8} constructed the YOLOv8-CML model for melon ripeness segmentation, which performed well under controlled conditions. However, like many models applied in agriculture, it encountered challenges in high-density planting environments. Occlusion and overlap of fruits in such scenarios made it difficult to maintain consistent recognition performance, underscoring the inherent difficulty in segmenting objects accurately when faced with overlapping targets and dense vegetation.
Sampurno et al. \citep{sampurno2024intrarow} deployed the YOLOv8n-seg model for robotic weeding, achieving notable segmentation accuracy. Yet, in complex field environments with dense weed growth, the model's performance was affected by the challenges of distinguishing between targets and their backgrounds. These results demonstrate the importance of balancing computational efficiency with the need for accurate segmentation in intricate and resource-constrained environments.
Wang et al. \citep{wang2023performance} applied the YOLOv8-seg model to a litchi picking robot, focusing on extracting regions of interest (ROIs) for litchi fruits and branches. Although the model contributed to efficient fruit-picking, occlusion and varying fruit sizes introduced challenges, particularly when overlapping occurred, which affected the precision of ROI extraction. This highlights the ongoing challenge of adapting segmentation models to dynamic agricultural scenarios involving variable object sizes.
Yue et al. \citep{yue2023improved} implemented SimConv in the YOLOv8-seg model for segmenting healthy and diseased tomato plants at different growth stages. Despite the improvements in segmentation accuracy, performance bottlenecks persisted in complex environments with significant occlusion and varying plant sizes. The model’s difficulties in such conditions reflect the broader challenges that arise when applying segmentation algorithms in agricultural contexts, where computational resources are often limited and crop characteristics can vary widely.

Although these studies have made progress in specific applications, challenges such as changing target sizes, complex ground backgrounds, and the correct identification of crop categories continue to make segmentation even more difficult. Ongoing optimization is essential to enhance model robustness and adaptability in these contexts. Moreover, the limited computing resources of UAVs pose additional difficulties in accomplishing complex tasks. In response, this paper proposes a lightweight and efficient segmentation model for Banana Plantations, ALSS-YOLO-Seg. The model integrates the Adaptive Lightweight Channel Split and Shuffling (ALSS) \citep{10680397} module and an efficient lightweight attention mechanism, Multi-Scale Channel Attention (MSCA) module. Experimental results demonstrate that this model shows strong potential for application in resource-constrained environments.
The main contributions of this paper are as follows:
\begin{enumerate}
\item This study uses SAM2 zero-shot capabilities to develop an iterative optimization annotation pipeline for segmentation mask generation, which significantly reduces data annotation costs and time. This pipeline leverages a small amount of weak annotations to automatically generate high-quality segmentation masks, minimizing annotation expenses and human intervention, thereby improving annotation efficiency and segmentation accuracy.

\item The ALSS-YOLO-Seg model is designed to integrate the ALSS module with the MSCA attention mechanism. The ALSS module optimizes feature extraction through an adaptive channel splitting strategy and enhances inter-channel information exchange via a channel shuffling mechanism, all while employing a bottleneck structure to reduce model complexity. This module helps to accurately identify crops. Furthermore, the MSCA functions as an efficient lightweight attention module, combining multi-scale feature extraction with channel attention. This combination significantly boosts the model's accuracy and generalization capabilities while maintaining low computational overhead, making the ALSS-YOLO-Seg model highly effective for tasks involving varying target scales, complex backgrounds, and resource-constrained scenarios.

\item We evaluated the zero-shot segmentation performance of SAM2 on the ADE20K \citep{zhou2019semantic} and Javeri \citep{Javeri-dataset} datasets. On the ADE20K dataset, SAM2-b, utilizing MBB-based prompts, achieved a segmentation performance of 75.7\% mIoU, surpassing the supervised BEIT-3\citep{wang2023image} model by 13.7\%. Utilizing the Javeri dataset, we further demonstrated that the iterative optimization annotation pipeline significantly reduces manual annotation workload while achieving high-quality data labeling with an mIoU of 93.8\%. Extensive experiments on our custom Banana Plantation segmentation dataset revealed that ALSS-YOLO-Seg (with a parameter count of 1.8256M) achieved state-of-the-art performance, with a \(mAP50_{mask}\), exceeding that of the second-best model, YOLOv8-Seg' (with 1.6952M parameters) by 1\%. Furthermore, significant improvements in both \(mAP50_{mask}\) scores and parameter efficiency were observed compared to other advanced segmentation models.
\end{enumerate}

\section{Materials and methods}

\subsection{Acquisition of images for banana plantations}

\begin{figure}[h!]
\begin{center}
\includegraphics[width=17cm]{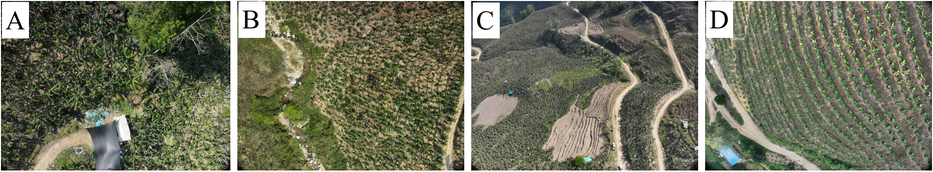}
\end{center}
\caption{ Captured banana plantation images obtained with the DJI Phantom 3M illustrate various scenarios, including: (A) a tree and house occlusion scene, (B) a complex scene with abundant weeds, and (C, D) variations in target sizes due to differing flight altitudes of the UAV, where (C) corresponds to a flight altitude of 12 meters and (D) corresponds to a flight altitude of 5 meters.}\label{Banana-Plantations}
\end{figure}

The images of Banana Plantations were acquired in Hekou Yao Autonomous County, Honghe Hani and Yi Autonomous Prefecture, Yunnan Province, China. The images were captured using a DJI Phantom 3M quadcopter UAV equipped with an RGB camera, which took vertical photographs of the ground at a shutter speed of 2 seconds to ensure clarity. To minimize image blur, the UAV hovered at each waypoint while capturing images. The forward and lateral overlap rates were set to 80\% and 90\%, respectively, and the images were saved in JPG format. Data collection took place daily from 10:00 AM to 12:00 PM between February and April 2024 to ensure consistent lighting conditions and image quality. To enhance the model's generalization capability for image segmentation of Banana Plantations in varying environments, images were captured at UAV flight heights of 5 meters, 8 meters, and 12 meters. Ultimately, a total of 3,880 raw images were obtained, covering various growth stages, heights, and different plots. Figure~\ref{Banana-Plantations} shows various scenarios.
% 目标尺寸变化、地形背景复杂、计算资源有限以及作物类别的正确识别等挑战使得分割更加困难。
In the subsequent sections, we will provide a detailed description of an iterative optimization process for generating segmentation masks from the raw images. Additionally, to validate the effectiveness of this pipeline, we utilized the ADE20K \citep{zhou2019semantic} and Javeri \citep{Javeri-dataset} datasets in our experiments.

\subsection{Iterative optimization annotation pipeline for segmentation mask generation}

\begin{figure}[h!]
\begin{center}
\includegraphics[width=17cm]{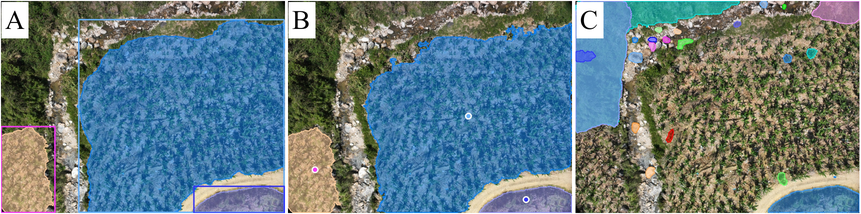}
\end{center}
\caption{ Comparison of segmentation results using the SAM2-b model with different prompting strategies: (A) Segmentation results with bounding box prompts; (B) Segmentation results with positive point prompts; (C) Segmentation results using the segment-anything mode without any prompts. The results demonstrate that the SAM2-b model achieves more accurate segmentation when guided by prompts, while the segment-anything mode alone fails to produce satisfactory segmentation results.}\label{SAM2-prompt}
\end{figure}

In the field of computer vision, recent segmentation models, such as the SAM2, have demonstrated remarkable performance across various tasks due to their strong generalization capabilities and cross-scenario segmentation effectiveness. However, the performance of these models heavily relies on high-quality user prompt, including positive points, negative points, and bounding boxes, as illustrated in Figure~\ref{SAM2-prompt}. This implies that, to generate accurate segmentation masks, users must manually annotate multiple key points or provide initial bounding boxes within the images to guide SAM2 in accurately locating target areas. Furthermore, traditional segmentation methods often require extensive finely annotated datasets, with the annotation process being time-consuming and labor-intensive. This is particularly challenging for large-scale agricultural scenarios involving UAV imagery, where annotation costs can be prohibitive.

\begin{figure}[h!]
\begin{center}
\includegraphics[width=17cm]{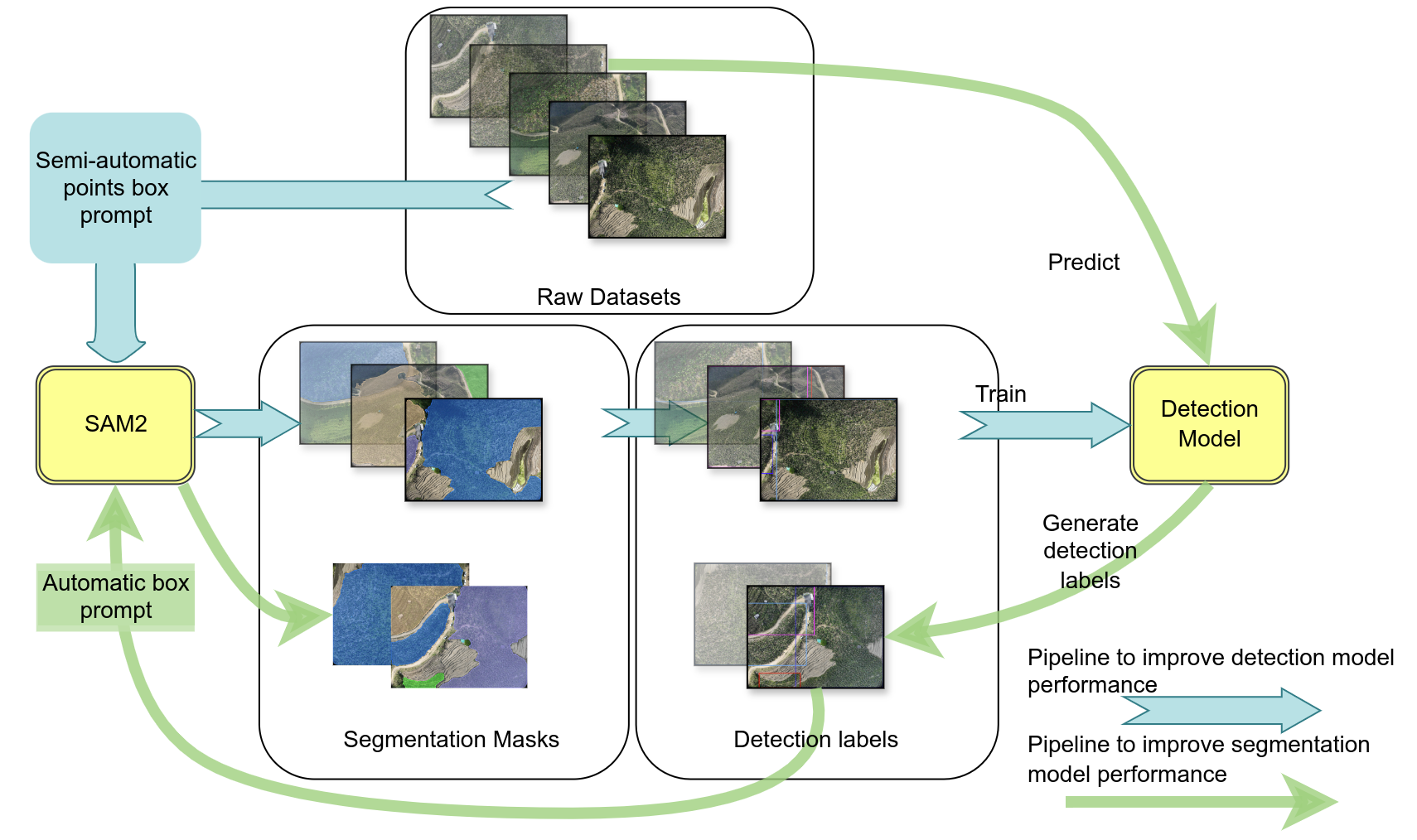}
\end{center}
\caption{ The proposed iterative optimization annotation pipeline for segmentation mask generation}\label{pipeline}
\end{figure}

To address these issues, we proposed an iterative optimization annotation pipeline for segmentation mask generation, aimed at reducing data annotation costs. The pipeline combines automated and semi-automated segmentation techniques to achieve precise delineation of target areas, significantly enhancing the quality of segmentation masks while minimizing manual intervention. The proposed pipeline is illustrated in Figure~\ref{pipeline}.

Initially, a randomly selected subset of images \( I_i\) from the extensive collection of banana plantation imagery was used for segmentation. Positive and negative hint points, denoted as \( P^+ = \{p_1^+, p_2^+, \dots, p_p^+\} \) and \( P^- = \{p_1^-, p_2^-, \dots, p_n^-\} \), along with bounding boxes \( B = \{b_1, b_2, \dots, b_b\} \), were employed as prompts to guide the segmentation process. These prompts provided critical information regarding target and non-target regions, allowing SAM2 to generate high-quality segmentation masks \( M_i = \{M_1, M_2, \dots, M_m\} \)  for each selected image.

\begin{equation}
    M_i = \text{SAM2}(I_i, P^+, P^-, B)
\end{equation}
Subsequently, we further processed the segmentation masks \( M_i \) by calculating the minimum bounding boxes (MBB), \( B_i \), for each mask, such that:

\begin{equation}
B_i = \left( \min(x), \min(y), \max(x), \max(y) \mid (x, y) \in M_i \right)
\end{equation}

This provides the initial location and extent of the target areas. The bounding boxes \( B_i \) serve as initial training data for a target detection model \( DET \), where the model learns the features of the targets from the labeled images and locates similar targets in the unlabeled images. After the initial training phase, the target detection model automatically generates preliminary bounding boxes \( B_j^{\text{pred}} \) in the unlabeled images \( I_j (j \notin i) \), significantly improving the efficiency and automation of target detection.
\begin{equation}
B_j^{\text{pred}}=DET(I_j)
\end{equation}
Next, the high-quality bounding boxes \( B_k^{\text{fine}} (k \in j) \) automatically generated by the object detection model were reintroduced into SAM2 as prompt information to generate accurate segmentation masks \( M_k^{\text{fine}} \).

\begin{equation}
    M_k^{\text{fine}} = \text{SAM2}(I_k, B_k^{\text{fine}})
\end{equation}
Using the newly generated refined mask, we further refine the MBB.
These new bounding boxes, along with their corresponding images, were then used to retrain the target detection model \( DET \), further improving its accuracy.

The process was repeated iteratively, with each iteration \( t \) progressively optimizing both the segmentation masks \( M_j^{(t)} \) and the bounding boxes \( B_j^{(t)} \). The collaborative interaction between the target detection model \( DET \) and SAM2 during each iteration resulted in increasingly accurate segmentation masks, ultimately achieving high-quality segmentation and detection of target areas within the images. This iterative process is modeled as:

\begin{equation}
M_j^{(t+1)} = \text{SAM2}(B_j^{(t)})
\end{equation}

\begin{equation}
B_j^{(t+1)} = \left( \min(x), \min(y), \max(x), \max(y) \mid (x, y) \in M_j^{(t+1)} \right)
\end{equation}

This iterative optimization process continues until convergence, which is determined when the difference between $B_j^{(t+1)}$ and $B_j^{t}$ is less than a predefined threshold $\epsilon$:

\begin{equation}
| B_j^{(t+1)} - B_j^{t} | < \epsilon 
\end{equation}
This optimization mechanism not only improves segmentation performance but also decreases dependency on manual prompts, thus significantly reducing data annotation costs.

We adopted targeted optimization strategies for the missed detection and false positive problems in a very small number of images after iteration. For missed detections (i.e., cases with insufficient recall), we manually provided initial prompt points to supplement the missing bounding boxes, ensuring that all targets are detected and segmented. For false detections (i.e., cases with insufficient precision), we adjusted the bounding boxes through appropriate displacement and scaling to correct the boundaries of the misidentified areas, thereby ensuring the accuracy of the segmentation masks.

This iterative optimization annotation pipeline possesses several advantages: first, it effectively leverages the powerful segmentation capabilities of the SAM2 model alongside the feature learning capacity of the target detection model to achieve precise segmentation and detection of targets within images. Secondly, by integrating initial prompt inputs with iterative optimization, the model gradually corrects errors and improves precision, ultimately generating high-quality segmentation masks. Moreover, this method exhibits strong generalization capabilities, allowing it to effectively handle segmentation tasks for banana plantations across varied scenarios.

In summary, the iterative optimization annotation pipeline for segmentation mask generation proposed in this study successfully combines automated and semi-automated segmentation. Through multiple rounds of optimization, it significantly enhances the precision and quality of the segmentation masks. Additionally, this pipeline reduces reliance on manual prompts, thereby considerably diminishing the costs and workload associated with data annotation, providing a refined segmentation solution for UAV-based analysis of banana plantations.

\subsection{ALSS-YOLO-Seg: A lightweight and efficient banana plantation segmentation model}
After constructing the Banana Plantation segmentation dataset using the method described in Section 3.2, we developed a dedicated segmentation model, ALSS-YOLO-Seg, specifically designed for this dataset.
\subsubsection{Model architecture}
The ALSS-YOLO model \citep{10680397} builds upon the strengths of the YOLO (You Only Look Once) architecture, which is renowned for its balance between speed and accuracy, solidifying its position as a seminal approach in the field of computer vision. YOLO's single-stage detection framework is particularly well-suited for applications requiring high real-time performance, making it widely adopted across various domains. Moreover, the incorporation of a Feature Pyramid Network (FPN) \citep{lin2017feature} in YOLO models enhances their ability to manage targets with significant scale variations by effectively capturing multi-scale features. This architecture ensures that YOLO-based models maintain high detection accuracy, even in the presence of targets of varying sizes within a scene, thereby providing robust and reliable detection performance.

The ALSS-YOLO-Seg model's architecture is specifically tailored to address the challenges of banana plantation segmentation tasks by modifying elements within the YOLO series. At its core, ALSS-YOLO-Seg incorporates the ALSS \citep{10680397} module, which reduces the overall complexity of the algorithm while enhancing its ability to tackle specific challenges. The ALSS module employs an adaptive channel splitting strategy to optimize feature extraction and incorporates a channel shuffling mechanism to enhance information exchange between channels. This enables the model to effectively capture occlusion features, thereby facilitating accurate crop identification.

Additionally, we innovatively developed the MSCA module. MSCA is an efficient and lightweight attention module inspired by the channel reduction and expansion techniques of SENet \citep{hu2018squeeze}, ensuring the model remains lightweight. This module combines multi-scale feature extraction with a channel attention mechanism, enhancing model accuracy and generalization capability while maintaining low computational overhead. MSCA demonstrates significant advantages in handling tasks with varying target scales and complex backgrounds, making it highly suitable for resource-constrained scenarios. 

\begin{figure}[h!]
\begin{center}
\includegraphics[width=17cm]{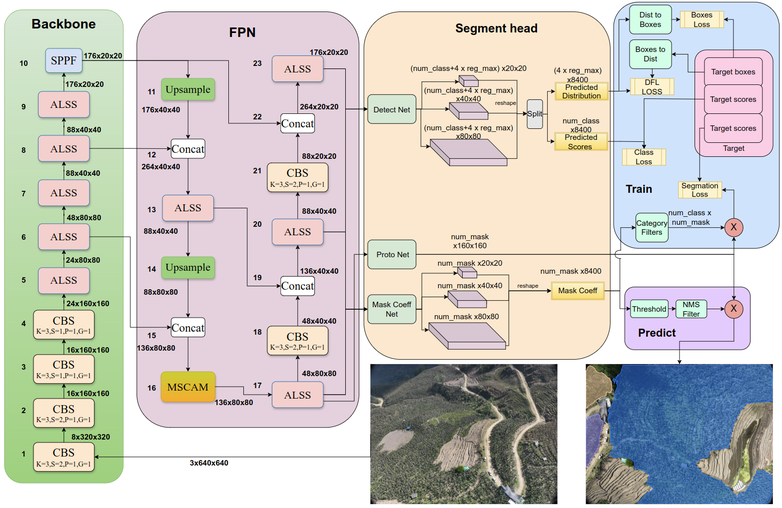}
\end{center}
\caption{ The architecture of the ALSS-YOLO-Seg segmenter. CBS denotes Convolution, Batch Normalization, and SiLU activation function. The symbol “k” represents the Kernel size, “s” denotes the Stride, and “p” indicates the Padding. }\label{ALSS-YOLO-Seg}
\end{figure}

The integration of these designs enables the ALSS-YOLO-Seg model to achieve higher performance and efficiency in the task of banana plantation segmentation. ALSS-YOLO-Seg leverages principles from YOLACT \citep{bolya2019yolact} for instance segmentation. It first extracts features from images using the backbone network and FPN, integrating features of different sizes. The output consists of detection and segmentation branches. The detection branch outputs class labels and bounding boxes, while the segmentation branch outputs k prototypes (default 32 in ALSS-YOLO-Seg) and k mask coefficients. The segmentation and detection tasks are computed in parallel. The segmentation branch inputs high-resolution feature maps, retaining spatial details and including semantic information. This map is processed through convolutional layers, upsampled, and then passed through two additional convolutional layers to output masks. The mask coefficients, similar to the classification branch of the detection head, range from -1 to 1. The instance segmentation results are obtained by multiplying and summing the mask coefficients with the prototypes.
Figure~\ref{ALSS-YOLO-Seg} illustrates the architecture of the ALSS-YOLO-Seg segmenter, and Table~\ref{param} outlines the primary parameters.

\begin{table*}[!h]
\centering
\caption{Parameter count and forward propagation runtime of the main modules in the ALSS-YOLO object detector.\label{param}}
\begin{tabular*}{\textwidth}{@{\extracolsep{\fill}}ccccc@{}}

% \begin{tabular}{ccccc}
\hline 
\textbf{Number}& \textbf{Module}&\textbf{Output}&\textbf{Params}&\textbf{GFLOPs}\\
\hline 
0 	& Input		& 3x640x640		& -			& -		    \\
1	& Conv		& 8x320x320		& 232		& 0.04		\\
2	& Conv		& 16x160x160	& 1184		& 0.06		\\
3	& Conv		& 16x160x160	& 2336		& 0.12		\\
4	& Conv 	    & 24x160x160	& 3504		& 0.18		\\
5	& ALSS  	& 24x80x80		& 2475		& 0.06		\\
6	& ALSS 	    & 48x80x80		& 3819		& 0.05		\\
7	& ALSS  	& 88x40x40		& 15020		& 0.07		\\
8	& ALSS 	    & 88x40x40		& 38393		& 0.12		\\
9	& ALSS		& 176x20x20		& 20886		& 0.03		\\
10	& SPPF	    & 176x20x20		& 77968		& 0.06		\\
11	& Upsample	& 176x40x40		& 0 		& 0.00		\\
12	& Concat 	& 264x40x40		& 0	        & 0.00		\\
13	& ALSS	    & 88x40x40		& 379477	& 1.21		\\
14	& Upsample	& 88x80x80		& 0			& 0.00		\\
15	& Concat	& 136x80x80		& 0	        & 0.00		\\
16	& MSCA 	& 136x80x80		& 7248		& 0.06		\\
17	& ALSS	    & 48x80x80		& 102963	& 1.31		\\
18	& Conv		& 48x40x40		& 20832		& 0.07		\\
19	& Concat	& 136x40x40		& 0			& 0.00		\\
20	& ALSS 	    & 88x40x40		& 68773	    & 0.22		\\
21	& Conv	    & 88x20x20		& 69872		& 0.06		\\
22	& Concat	& 264x20x20		& 0		    & 0.00		\\
23	& ALSS	    & 176x20x20		& 94872		& 0.08		\\
% \hline 
% \end{tabular}
% \end{table}

% \begin{table}[!t]
% \begin{tabular}{ccccc}
% \hline 
% \textbf{Number}& \textbf{Module}&\textbf{Output}&\textbf{Params}&\textbf{time(ms)}\\
% \hline 
24	& Segment 	& -		        & 918811	    & 5.59		\\
-	& Total		& -				& 1828665*	    & 9.39		\\
\hline 
\end{tabular*}
\noindent{\footnotesize{* The number of parameters following layer fusion is 1825589.}}
\end{table*}

\subsubsection{MSCA module}

Recent advancements in attention mechanisms have gained widespread application in computer vision, demonstrating their immense potential in enhancing feature extraction capabilities and optimizing computational efficiency. Channel attention mechanisms, in particular, have emerged as key components in numerous deep learning architectures, offering the advantage of selectively enhancing important features while suppressing irrelevant information. Prominent attention modules, such as SENet \citep{hu2018squeeze}, CBAM \citep{woo2018cbam}, and ECA-Net \citep{wang2020eca}, leverage distinct approaches to re-weight features. SENet pioneered the use of channel compression and expansion to refine feature representations, while CBAM integrated both spatial and channel attention to strengthen feature detection. ECA-Net introduced an efficient channel interaction method that reduces computational complexity, making it ideal for lightweight models. Moreover, the multi-scale feature extraction strategy employed in the Inception series networks \citep{szegedy2015going}\citep{ioffe2015batch}\citep{szegedy2016rethinking}\citep{szegedy2017inception} has been influential, demonstrating that capturing features at multiple scales can significantly improve a model's ability to understand complex scenes.

\begin{figure}[h!]
\begin{center}
\includegraphics[width=17cm]{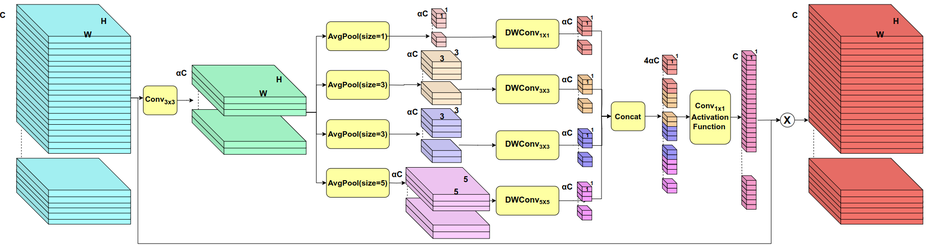}
\end{center}
\caption{ MSCA module structure diagram. }\label{MSCA}
\end{figure}

Inspired by SENet's channel compression and expansion techniques, as well as Inception's multi-scale design, this paper presents MSCA, a module designed to enhance feature representation in the task of banana plantation segmentation by combining multi-scale feature extraction with channel attention mechanisms. MSCA improves segmentation performance while maintaining computational efficiency through fine-tuned channel processing and multi-scale information fusion. Its internal structural details are illustrated in Figure ~\ref{MSCA}.

Initially, given an input feature map \(X \in \mathbb{R}^{H \times W \times C}\), a channel reduction layer employs a \(3 \times 3\) convolution to compress the input features from the original channel size \(C\) to a reduced channel size \(\alpha C\), where \(\alpha\) is the compression factor, typically set to \(1/32\):
\begin{equation}
Z \in \mathbb{R}^{H \times W \times \alpha C} = Conv_{3\times3}(X)
\end{equation}
This channel compression step significantly lowers computational complexity while emphasizing key features.

At its core, MSCA employs a multi-scale feature extraction mechanism, utilizing various convolution kernel sizes and pooling operations to capture features at different scales, thereby enhancing its ability to distinguish objects of varying sizes. To further optimize computational efficiency, the module incorporates depthwise convolutions. This design significantly reduces the computational burden while preserving essential correlations between feature channels. Following the advice of Ref.\cite{Depthwise}, the activation function is not employed following the depthwise convolution. We demonstrate the effectiveness of this approach in subsequent ablation experiments. Specifically, the module implements four sets of pooling and convolution operations without padding:

1. The first set employs global average pooling with a target output size of \(1 \times 1\) followed by a \(1 \times 1\) convolution:
\begin{equation}
    G_{1} \in \mathbb{R}^{1 \times 1 \times \alpha C} = DWConv_\text{1x1}(AvgPool(Z, \text{size}=1))
\end{equation}
    
2. The second set utilizes adaptive average pooling with a target output size of \(3 \times 3\) combined with a \(3 \times 3\) convolution:
\begin{equation}
    G_{2} \in \mathbb{R}^{1 \times 1 \times \alpha C} = DWConv_\text{3x3}(AvgPool(Z, \text{size}=3))
\end{equation}
    
3. The third set also employs adaptive average pooling with a target output size of \(3 \times 3\) and a \(3 \times 3\) convolution to capture more extensive local features:
\begin{equation}
    G_{3} \in \mathbb{R}^{1 \times 1 \times \alpha C} = DWConv_\text{3x3}(AvgPool(Z, \text{size}=3))
\end{equation}

4. The fourth set leverages adaptive average pooling with a target output size of \(5 \times 5\) in conjunction with a \(5 \times 5\) convolution:
\begin{equation}
    G_{4} \in \mathbb{R}^{1 \times 1 \times \alpha C} = DWConv_\text{5x5}(AvgPool(Z, \text{size}=5))
\end{equation}

After extracting multi-scale features through the four parallel convolutions, the resulting feature vectors are concatenated along the channel dimension to form a comprehensive multi-scale representation:
\begin{equation}
G_{concat} \in \mathbb{R}^{1 \times 1 \times 4\alpha C} = Concat(G_{1}, G_{2}, G_{3}, G_{4})
\end{equation}
This concatenated feature vector is then passed through a \(1 \times 1\) convolution for channel expansion, followed by the \(\sigma\) nonlinear activation function to generate the attention weights:
\begin{equation}
H \in \mathbb{R}^{1 \times 1 \times C} = \sigma(Conv_\text{1x1}(G_{concat}))
\end{equation}
These attention weights dynamically adjust the importance of each feature channel, thereby re-weighting the input features \(X\) accordingly:
\begin{equation}
Y \in \mathbb{R}^{H \times W \times C} = X \cdot H
\end{equation}
where \(Y\) represents the modulated feature map. After passing through the channel attention mechanism, the module enhances or suppresses specific feature channels, allowing the model to focus on the most relevant features while filtering out noise or irrelevant information. The introduction of this attention mechanism enables MSCA to selectively enhance critical features, significantly improving its ability to detect and segment multi-scale objects within the plantation segmentation task.

The MSCA module demonstrates significant versatility in tasks requiring precise feature extraction and multi-scale processing. Its lightweight architecture and efficient computational management enhance the model's ability to differentiate between objects of varying scales in segmentation tasks. By integrating multi-scale feature extraction with channel attention mechanisms, MSCA substantially improves feature representation and performance. This resource-efficient design makes the module particularly well-suited for deployment in resource-constrained environments, such as mobile devices and UAV platforms, thereby facilitating real-time segmentation capabilities. Consequently, MSCA offers a robust solution for complex tasks, including applications like banana plantation segmentation.

\subsection{Experiment setup}
All experiments requiring model training were conducted with convergence achieved after 200 epochs. The input image size was set to 640×640 pixels, and the Stochastic Gradient Descent (SGD) optimizer was used with the following parameters: a batch size of 120, momentum of 0.937, and weight decay of 0.0005. To enhance the stability of the training process, a 3-epoch warm-up phase was employed, during which the optimizer's momentum was initialized at 0.8. After the warm-up phase, the learning rate was progressively decayed using a cosine annealing schedule, starting from an initial value of 0.001 and gradually decreasing to a minimum of 0.00001.Instead of applying offline data augmentation to the original images, we implemented a series of online augmentation techniques, including mosaic augmentation, random perturbations (such as rotation and scaling), mixing, color perturbation, and random flipping. These techniques were employed to enhance the model's generalization ability.

The experiments were conducted on a machine running the Ubuntu 20.04 LTS operating system, equipped with an AMD EPYC 7543 32-Core Processor, an NVIDIA GeForce RTX 3090 GPU with 24GB of memory. The software environment included Python 3.9.19, PyTorch 2.4.1, and CUDA 12.0.

\subsection{Evaluation metrics}

In this study, we evaluate two output types: bounding box and mask predictions. The following metrics are used to assess model performance: box precision (\(P_{box}\)), box recall (\(R_{box}\)), box F1 score (\(F1_{box}\)), box average precision (\(AP_{box}\)), box mean average precision (\(mAP_{box}\)), mask precision (\(P_{mask}\)), mask recall (\(R_{mask}\)), mask F1 score (\(F1_{mask}\)), mask average precision (\(AP_{mask}\)), mask mean average precision (\(mAP_{mask}\)), mean Intersection over Union (mIoU), frame rate (FPS), and the number of model parameters. Unless stated otherwise, precision and recall are calculated at an IoU threshold of 0.5 and a confidence threshold corresponding to the maximum F1 score.

Box metrics evaluate detection performance: \(P_{box}\) measures the proportion of correctly predicted boxes, \(R_{box}\) captures the model's ability to detect all relevant objects, \(F1_{box}\) provides a harmonic mean of \(P_{box}\) and \(R_{box}\), and \(AP_{box}\) and \(mAP_{box}\) reflect the precision-recall trade-off and average performance across categories. The formulas are as follows:

\begin{equation}
P_{box} = \frac{TP_{box}}{TP_{box} + FP_{box}}
\end{equation}
\begin{equation}
R_{box} = \frac{TP_{box}}{TP_{box} + FN_{box}}
\end{equation}
\begin{equation}
F1_{box} = \frac{2 \cdot P_{box} \cdot R_{box}}{P_{box} + R_{box}}
\end{equation}
\begin{equation}
AP_{box} = \int_0^1 P_{box}(R_{box}) dR_{box}
\end{equation}
\begin{equation}
mAP_{box} = \frac{1}{n} \sum_{i=1}^{n} AP_{box, i}
\end{equation}

Mask metrics assess semantic segmentation quality: \(P_{mask}\) and \(R_{mask}\) measure pixel-level accuracy and completeness, \(F1_{mask}\) balances the two, and \(AP_{mask}\) and \(mAP_{mask}\) evaluate performance across categories. The mIoU metric captures the overlap between predicted and ground truth regions. The formulas are:

\begin{equation}
P_{mask} = \frac{TP_{mask}}{TP_{mask} + FP_{mask}}
\end{equation}
\begin{equation}
R_{mask} = \frac{TP_{mask}}{TP_{mask} + FN_{mask}}
\end{equation}
\begin{equation}
F1_{mask} = \frac{2 \cdot P_{mask} \cdot R_{mask}}{P_{mask} + R_{mask}}
\end{equation}
\begin{equation}
AP_{mask} = \int_0^1 P_{mask}(R_{mask}) dR_{mask}
\end{equation}
\begin{equation}
mAP_{mask} = \frac{1}{n} \sum_{i=1}^{n} AP_{mask, i}
\end{equation}
\begin{equation}
mIoU = \frac{1}{n} \sum_{i=1}^{n} \frac{|A_{pred} \cap A_{gt}|}{|A_{pred} \cup A_{gt}|}
\end{equation}

Here, \(TP\), \(FP\), and \(FN\) denote true positives, false positives, and false negatives, respectively, and \(n\) represents the number of categories. \(A_{pred}\) and \(A_{gt}\) refer to the predicted and ground truth areas for mIoU calculation.

\section{Results and discussion}

In this section, we first conduct an experimental evaluation of prompt-based zero-shot instance segmentation on the ADE20K \citep{zhou2019semantic} and Javeri \citep{Javeri-dataset} datasets. Following this, we validate the iterative optimization process for generating segmentation masks on the Javeri dataset, demonstrating that the method produces high-quality segmentation outputs. Additionally, we test our ALSS-YOLO-Seg segmenter on custom Banana Plantation segmentation dataset, showcasing its state-of-the-art performance.

\subsection{Experimental evaluation of prompt-based Zero-Shot instance segmentation on the ADE20K dataset}

\begin{figure}[h!]
\begin{center}
\includegraphics[width=17cm]{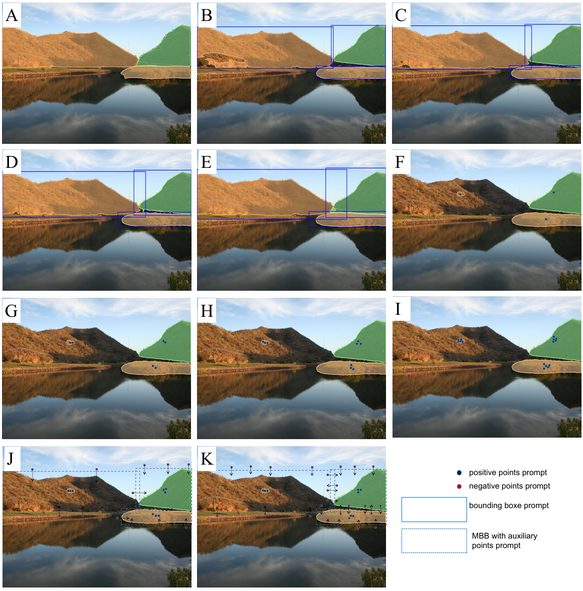}
\end{center}
\caption{ Schematic Illustration of Prompt Configurations and Segmentation Results for SAM2\_b; (A) ground truth; (B) MBB prompts; (C) MBB+5\% prompts; (D) MBB+10\% prompts; (E) 1 Ppoint prompts; (F) 2 Ppoints prompts; (G) 3 Ppoints prompts; (H) 5 Ppoints prompts; (J) 3 Ppoints 4 Npoints prompts; (K) 3 Ppoints 8 Npoints prompts.}\label{prompts_schematic}
\end{figure}

\begin{figure}[h!]
\begin{center}
\includegraphics[width=17cm]{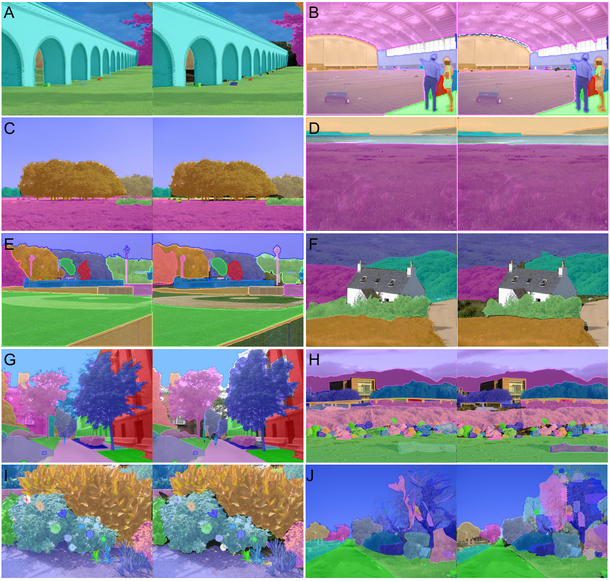}
\end{center}
\caption{ Comparison of MBB prompt segmentation results with ground truth: A total of 10 comparative pairs are presented, with ground truth images displayed on the left and MBB prompt segmentation from the SAM2\_b model on the right}\label{ADE20K-Figure}
\end{figure}

To evaluate the zero-shot segmentation performance of the SAM \citep{kirillov2023segment} and SAM2 \citep{ravi2024sam} models under different parameter configurations, we utilized the mIoU metric on the ADE20K validation set, which contains 2,000 images with detailed annotations across 150 densely predicted categories. This provides a rich and diverse dataset for model performance validation. The experimental prompts include the use of the Minimum Bounding Box (MBB) and its variants with boundary expansions, as well as point-based prompts. The MBB represents the smallest rectangle that encloses the ground truth bounding box in the ADE20K annotations. Variants such as MBB+5\%, MBB+10\%, and MBB+20\% correspond to increasing the width and height of the original bounding box by 5\%, 10\%, and 20\%, respectively. These variations simulate the practical challenges of achieving precise manual annotation of the minimum bounding box in densely annotated scenes. In addition to bounding boxes, we incorporated point-based prompts: "1 Ppoint" denotes a single positive point prompt, while "2 Ppoints," "3 Ppoints," and "5 Ppoints" refer to 2, 3, and 5 positive point prompts, respectively. Similarly, "3 Ppoints 4 Npoints" and "3 Ppoints 8 Npoints" represent 3 positive point prompts combined with 4 or 8 negative point prompts. A schematic diagram of the prompt is shown in Figure~\ref{prompts_schematic}. The experimental results are shown in Table~\ref{ADE20K}. Additionally, Table~\ref{TIME} summarizes the processing time for these images by the SAM model and provides statistics on model parameter sizes. Figure~\ref{ADE20K-Figure} further illustrates a comparison between the segmentation results of the MBB prompt and the ground truth.

% prompts_schematic

As shown in Figure~\ref{prompts_schematic}, the point-based prompt segmentation results are generally inferior to the box-based prompt segmentation results. When using points as prompts, the model generates three levels of hierarchical masks: whole, part, and sub-part. The output is determined based on confidence ranking, where the mask with the highest confidence is selected. However, in some cases, we aim to focus on the whole object, but the segmentation result may return a sub-part mask. In such cases, additional prompts are often required to achieve the desired segmentation outcome.

\begin{table*}[!h]
\centering
\caption{The mIoU results of segmentation for the SAM and SAM2 models using various prompts, assuming ADE20K validation set annotations as ground-truth. \label{ADE20K}}
\resizebox{\textwidth}{!}{ % Resize the table to fit the text width
\begin{tabular}{cccccccc}
\hline
\textbf{Prompt} & \textbf{mobile\_SAM} & \textbf{SAM\_b} & \textbf{SAM\_l} & \textbf{SAM2\_t} & \textbf{SAM2\_s} & \textbf{SAM2\_b} & \textbf{SAM2\_l} \\
\hline
MBB         & 0.7366 & 0.7307 & 0.7452 & 0.7535 & 0.7529 & 0.7566 & 0.7556 \\
MBB+5\%     & 0.7332 & 0.7295 & 0.7440 & 0.7504 & 0.7498 & 0.7536 & 0.7524 \\
MBB+10\%    & 0.7272 & 0.7249 & 0.7408 & 0.7438 & 0.7437 & 0.7474 & 0.7462 \\
MBB+20\%    & 0.7103 & 0.7096 & 0.7289 & 0.7222 & 0.7203 & 0.7260 & 0.7254 \\
1 Ppoint     & 0.4830 & 0.4805 & 0.4981 & 0.5162 & 0.5164 & 0.5323 & 0.5340 \\
2 Ppoints     & 0.4918 & 0.4907 & 0.5088 & 0.5162 & 0.5164 & 0.5322 & 0.5340 \\
3 Ppoints     & 0.4957 & 0.4993 & 0.5153 & 0.5162 & 0.5164 & 0.5323 & 0.5340 \\
5 Ppoints     & 0.4790 & 0.5052 & 0.5211 & 0.5162 & 0.5164 & 0.5322 & 0.5340 \\
3 Ppoints 4 Npoints   & 0.5320 & 0.5539 & 0.5167 & 0.5162 & 0.5164 & 0.5323 & 0.5340 \\
3 Ppoints 8 Npoints   & 0.5267 & 0.5507 & 0.5151 & 0.5162 & 0.5164 & 0.5322 & 0.5340 \\
\hline
\end{tabular}}
\end{table*}

\begin{table*}[!h]
\centering
\caption{Processing Time and Model Parameter Statistics for SAM and SAM2 Models.\label{TIME}}
\resizebox{\textwidth}{!}{ % Resize the table to fit the text width
\begin{tabular}{cccccccc}
\hline
\textbf{} & \textbf{mobile\_SAM} & \textbf{SAM\_b} & \textbf{SAM\_l} & \textbf{SAM2\_t} & \textbf{SAM2\_s} & \textbf{SAM2\_b} & \textbf{SAM2\_l} \\
\hline
Parameters  & 10.1(M) & 93.7(M) & 312.3(M) & 38.9(M) & 46.0(M) & 80.8(M) & 224.4(M) \\
Total time* & 1876(S) & 6482(S) & 14643(S) & 3142(S) & 3434(S) & 5080(S) & 9987(S) \\
\hline
\end{tabular}}
\noindent{\footnotesize{* The total time includes the combined inference time for the image encoder, prompt encoder, and mask decoder when processing 2,000 images from the ADE20K dataset. Inference was performed using a single NVIDIA RTX 3090 GPU.}}
\end{table*}

Table~\ref{ADE20K} illustrates that the segmentation performance is heavily impacted by varying prompt configurations, with SAM2 models consistently surpassing SAM models in the majority of cases.
Firstly, the results for the MBB and its extended variants demonstrate that both SAM and SAM2 models achieve their best segmentation performance with the standard MBB prompt. Notably, the larger models, such as SAM2\_l and SAM2\_b, achieve mIoU scores of 0.7556 and 0.7566, respectively, under the standard MBB prompt, highlighting their superior segmentation capability. However, as the bounding box is progressively expanded, the segmentation performance of all models exhibits a noticeable decline. This trend is particularly evident with the MBB+20\% prompt, where all models show a marked decrease in mIoU, suggesting that excessive expansion of the bounding box introduces more background noise, which in turn diminishes the model's segmentation precision. For instance, mobile\_sam sees its mIoU drop from 0.7366 with the MBB prompt to 0.7103 with the MBB+20\% prompt, reflecting its sensitivity to boundary extensions.
Secondly, the performance of point-based prompts is generally lower compared to the MBB prompts. With a single positive point prompt (1 Ppoint), the mIoU scores of all models drop considerably, indicating that a single point does not provide sufficient spatial information to guide effective segmentation. While increasing the number of positive point prompts (e.g., 2 Ppoints, 3 Ppoints, and 5 Ppoints) leads to some improvement in segmentation performance, the gains remain limited. For example, SAM2\_l shows identical mIoU values of 0.5340 for both 1 Ppoint and 3 Ppoints prompts, suggesting that increasing the number of positive points does not significantly enhance segmentation performance.
The incorporation of both positive and negative point prompts can improve segmentation results. For instance, mobile\_SAM shows an mIoU increase from 0.4957 with 3 Ppoints to 0.5320 with the 3 Ppoints 4 Npoints prompt, while SAM\_b improves from 0.4993 to 0.5539 under the same configuration. The inclusion of negative points helps to exclude incorrect regions, thereby refining the object boundaries and enhancing segmentation accuracy. However, despite these improvements, the performance of point-based prompts still lags behind that of the MBB prompts, underscoring the limitations of point-based prompts in providing sufficient spatial context.
% Overall, SAM2 models outperform SAM models across all prompt configurations, with larger SAM2 models such as SAM2\_l and SAM2\_b demonstrating superior generalization ability and segmentation accuracy in nearly every case. These results indicate that the structural enhancements and parameter optimizations in the SAM2 models contribute to their improved segmentation performance.

Based on the results shown in Table~\ref{ADE20K} and ~\ref{TIME}, there is a clear trend that model performance, as measured by mIoU, generally improves with an increase in the number of parameters. Larger models such as SAM2\_l and SAM2\_b outperform their smaller counterparts in both the SAM and SAM2 series. However, a significant observation is that even the smaller variants of SAM2, like SAM2\_t and SAM2\_s, achieve better generalization performance on the ADE20K validation set compared to SAM models with larger parameter counts.
This improvement can be attributed to the SAM2 models being trained on a larger and more diverse dataset(SA-V dataset). The SA-V dataset contains significantly more diverse annotations and richer content, giving the SAM2 model stronger Zero-Shot capabilities. As a result, despite their smaller size, SAM2 models like SAM2\_t and SAM2\_s outperform larger SAM models due to the advantage provided by the extensive training on the SA-V dataset. This highlights the importance of the training data’s quality and diversity in improving segmentation performance.

% prompts_schematic

Figure~\ref{ADE20K-Figure} exhibits that SAM2\_b, when prompted with MBB, performs suboptimally in segmenting elongated objects compared to the ground truth provided by the ADE20K dataset. For instance, in Figure~\ref{ADE20K-Figure}B, the segmentation of the shoulder bag is inferior to the ground truth. Additionally, SAM2\_b tends to segment local areas, such as the football field in Figure~\ref{ADE20K-Figure}E. However, we also observed that SAM2\_b produces smoother segmentation boundaries compared to the sharper edges in the ground truth (Figures~\ref{ADE20K-Figure}F, ~\ref{ADE20K-Figure}G, ~\ref{ADE20K-Figure}H, and ~\ref{ADE20K-Figure}J). Furthermore, when the quality of the ground truth is poor, the SAM2\_b model, even with MBB prompting, struggles to produce satisfactory results, as illustrated in Figure~\ref{ADE20K-Figure}K.

\begin{table*}[!h]
\centering
\caption{Comparison of Model Performance on ADE20K Validation Set.\label{ADE20K_Validation}}
\begin{tabular}{ccccccc}
\hline
\textbf{Model} & \textbf{train set}& \textbf{Prompt} & \textbf{Crop Size} & \textbf{mIoU}\\
\hline
HorNet\citep{rao2022hornet}      & ADE20K train set & No & 640$^2$ & 0.575\\
SeMask\citep{jain2023semask}      & ADE20K train set & No & 640$^2$ & 0.570\\
SwinV2-G\citep{liu2022swin}    & ADE20K train set & No & 896$^2$ & 0.593\\
ViT-Adapter\citep{chen2022vision} & ADE20K train set & No & 896$^2$ & 0.594\\
Mask DINO\citep{li2023mask}   & ADE20K train set & No & -       & 0.595\\
BEIT-3\citep{wang2023image}      & ADE20K train set & No & 896$^2$ & 0.620\\
SAM2\_b\citep{ravi2024sam2}     & SA-V             & MBB & 1024$^2$ & 0.757\\
SAM2\_l\citep{ravi2024sam2}     & SA-V             & MBB & 1024$^2$ & 0.756\\
\hline
\end{tabular}
\end{table*}

In Table \ref{ADE20K_Validation}, we compare the segmentation performance of various models on the ADE20K validation set. It is important to highlight that while most models were trained specifically on the ADE20K training set, the SAM2 models, both SAM2\_b and SAM2\_l, achieved their results in a zero-shot manner, without any training on ADE20K. Instead, SAM2 models utilized MBB prompts during inference. Despite not being fine-tuned on the dataset, SAM2 demonstrates superior segmentation performance, with mIoU scores of 0.757 and 0.756 for SAM2\_b and SAM2\_l, respectively. These results significantly surpass the best-performing trained models, such as BEIT-3 (0.620 mIoU). This outcome highlights the exceptional generalization capabilities of SAM2, showing that even without direct training, it can achieve state-of-the-art results by leveraging effective prompts.

\subsection{Experimental evaluation of prompt-based zero-shot instance segmentation on the Javeri dataset}

To further assess the generalization capability of the SAM model in UAV-based plantation scenarios, we employed the Javeri dataset, which is specifically tailored for segmentation tasks in plantations captured from UAV perspectives. The dataset consists of 280 training images and 40 validation images, all meticulously annotated. We selected mIoU as the evaluation metric to assess the zero-shot instance segmentation capabilities of both the SAM and SAM2 models. The experimental results are presented in Table~\ref{Javeri}.

\begin{table*}[!h]
\centering
\caption{The mIoU results of segmentation for the SAM and SAM2 models using various prompts, assuming Javeri validation set annotations as ground-truth. \label{Javeri}}
\resizebox{\textwidth}{!}{ % Resize the table to fit the text width
\begin{tabular}{cccccccc}
\hline
\textbf{Prompt} & \textbf{mobile\_SAM} & \textbf{SAM\_b} & \textbf{SAM\_l} & \textbf{SAM2\_t} & \textbf{SAM2\_s} & \textbf{SAM2\_b} & \textbf{SAM2\_l} \\
\hline
MBB         & 0.9622 & 0.9663 & 0.9654 & 0.9664 & 0.9691 & 0.9684 & 0.9680 \\
MBB+5\%     & 0.9602 & 0.9647 & 0.9647 & 0.9644 & 0.9672 & 0.9669 & 0.9665 \\
MBB+10\%    & 0.9581 & 0.9640 & 0.9633 & 0.9608 & 0.9651 & 0.9640 & 0.9632 \\
MBB+20\%    & 0.9417 & 0.9530 & 0.9481 & 0.9370 & 0.9518 & 0.9531 & 0.9386 \\
1 Ppoint    & 0.9511(-2) & 0.9513(-2) & 0.9574(-5) & 0.9580(-5) & 0.9534(-4) & 0.9611(-5) & 0.9594(-2) \\
2 Ppoints     & 0.9466(-5) & 0.9521(-2) & 0.9539(-6) & 0.9580(-5) & 0.9534(-4) & 0.9611(-5) & 0.9594(-2) \\
3 Ppoints     & 0.9469(-6) & 0.9524(-2) & 0.9519(-7) & 0.9580(-2) & 0.9534(-4) & 0.9611(-5) & 0.9594(-2) \\
\hline
\end{tabular}}
\noindent{\footnotesize{Only segmentation results with mIoU greater than 0.75 are reported, and the number in brackets indicates the number of images with mIoU lower than 0.75 compared to the ground truth.}}
\end{table*}

The results in Table \ref{Javeri} indicate that both the SAM and SAM2 models achieve high mIoU scores across different prompts, showcasing strong zero-shot instance segmentation performance on the plantation dataset. Overall, the SAM2 model consistently outperforms SAM, particularly with the MBB prompt, where it achieves the highest mIoU values. However, point-based prompts yield comparatively lower segmentation accuracy. This suggests that point-based prompts introduce ambiguity, consistent with the observations discussed in Section 3.1.

\subsection{Validation of the iterative optimization annotation pipeline}

\begin{figure}[h!]
\begin{center}
\includegraphics[width=12cm]{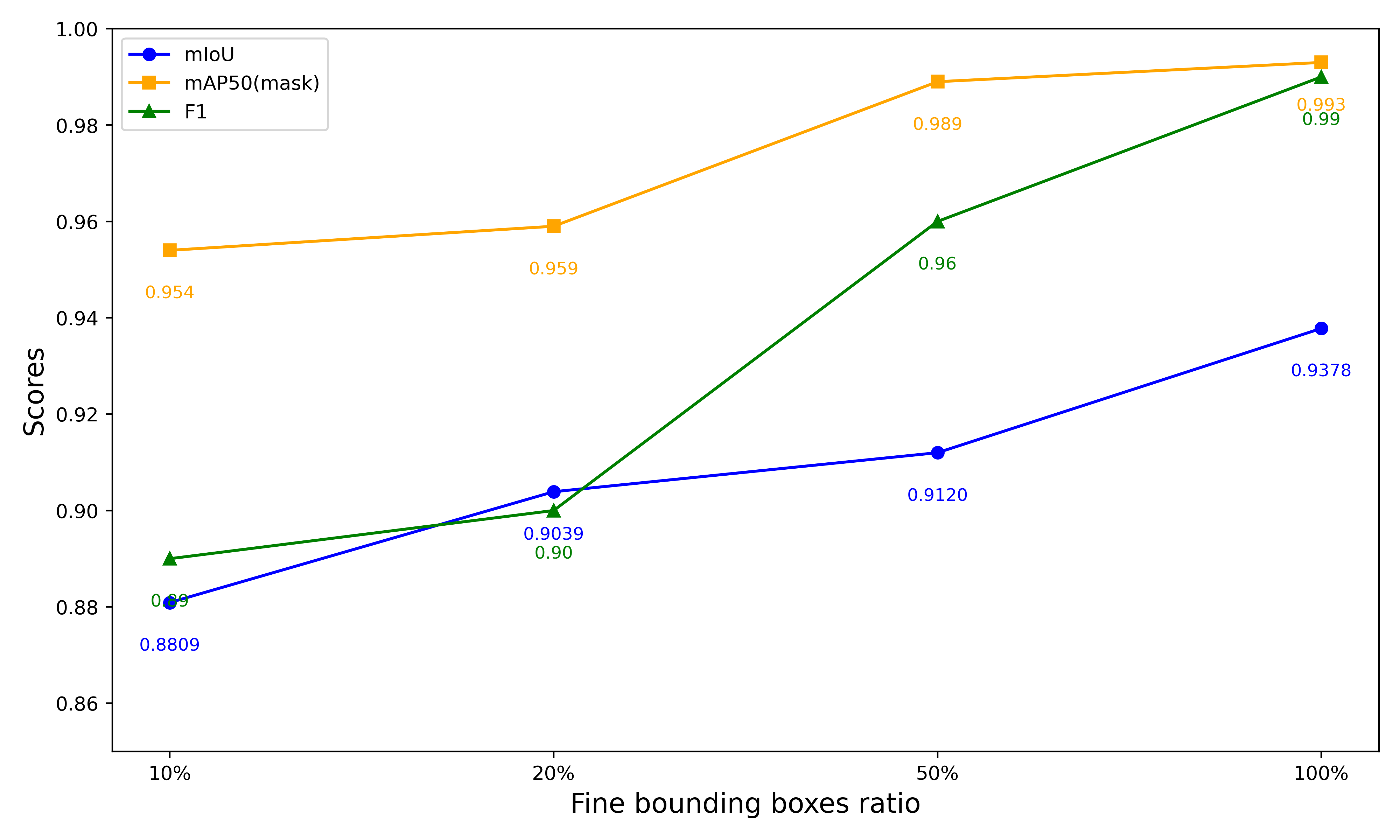}
\end{center}
\caption{Segmentation performance on the Javeri validation set at varying proportions of fine bounding boxes. }\label{Javeri-score}
\end{figure}

\begin{figure}[h!]
\begin{center}
\includegraphics[width=17cm]{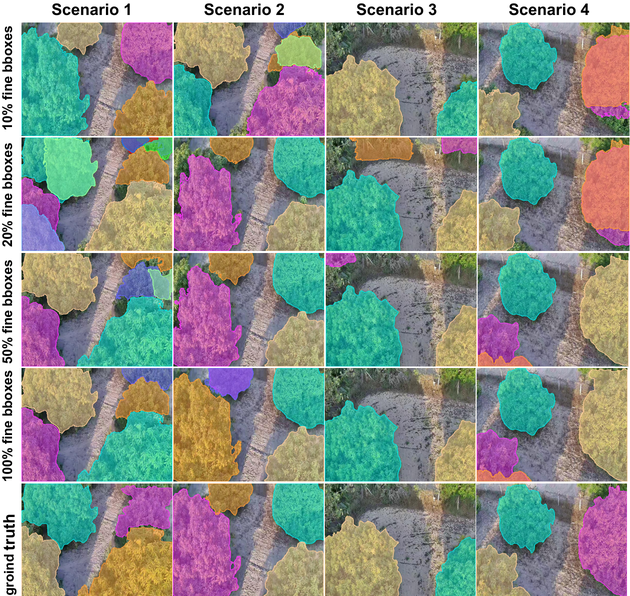}
\end{center}
\caption{ Qualitative analysis of segmentation performance with varying percentages of fine bounding boxes on the Javeri validation set.
}\label{Javeriv2}
\end{figure}

We employed the iterative optimization annotation pipeline described in Section 2.2 to generate segmentation masks and conducted experiments on the Javeri dataset. Specifically, we utilized YOLOv8-x as the object detector and SAM2\_b as the segmenter. In our experiments, we recorded the segmentation performance of YOLOv8-x on the Javeri validation set when the fine bounding boxes reached 10\%, 20\%, 50\% and 100\% respectively. The evaluation metrics included \( mAP50_{\text{mask}} \), F1 Score, and mIoU. The experimental results are presented in Figure~\ref{Javeri-score}. Figure~\ref{Javeriv2} qualitatively illustrates the impact of fine bounding boxes on segmentation performance.

% prompts_schematic

As shown in Figure~\ref{Javeri-score}, increasing the amount of annotated data significantly enhances the performance of segmentation, with the most noticeable improvements occurring when the amount of annotated data is relatively small. Specifically, when the proportion of annotated data increases from 10\% to 50\%, the performance gain is the most substantial. This indicates that the initial increase in annotated data provides the model with more critical target features, enabling it to better segment objects. This trend can also be observed in Figure~\ref{Javeriv2}, where the increase in fine bounding boxes leads to progressively higher-quality segmentation results. Additionally, we note that when the fine bounding boxes reach 100\%, the segmentation results of the SAM2\_b model are comparable to the ground truth. 
The results of this experiment demonstrate that the iterative optimization process is capable of generating high-quality segmentation annotations. Since the entire process only requires a small amount of manual annotation (and the manual annotation process only requires positive points, negative points and bounding boxes), this approach significantly reduces the cost of manual annotation compared to conventional methods.

\subsection{Evaluation of ALSS-YOLO-Seg on a custom Banana Plantation segmentation dataset.}
\begin{figure}[h!]
\begin{center}
\includegraphics[width=17cm]{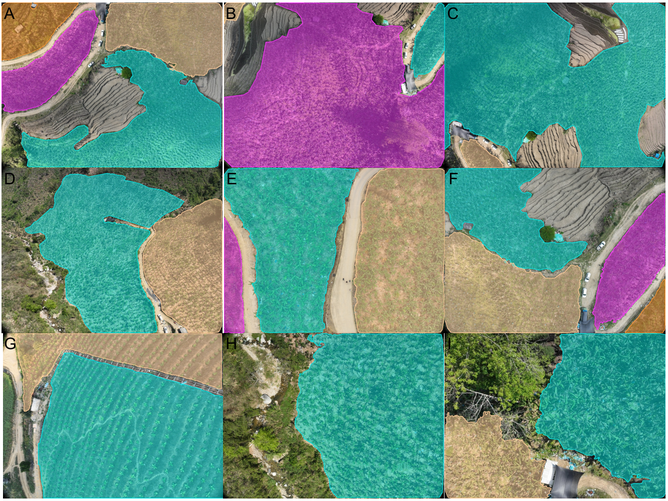}
\end{center}
\caption{ Annotated examples from the Banana Plantation segmentation dataset.}\label{banana}
\end{figure}

We constructed a meticulously annotated segmentation dataset for Banana Plantations using the iterative optimization annotation pipeline. Details of the original data are provided in Section 2.1. 
The processed dataset consists of 3,880 original images and their corresponding segmentation masks, divided into 3,492 pairs for the training set and 388 pairs for the test set.
Figure~\ref{banana} illustrates several annotated examples from the dataset, showcasing complex scenarios. These include occlusion caused by the banana transportation cable way (Figure~\ref{banana}G), interference from weeds (Figure~\ref{banana}H), and challenges arising from large target sizes and complex ground backgrounds in low-altitude UAV flight images. (Figure~\ref{banana}I). To address the practical deployment needs of UAV platforms, we employed this dataset to train a lightweight and efficient segmentation model, ALSS-YOLO-Seg. Extensive ablation studies and comparative experiments were carried out to validate the model's optimal performance.
% Segmentation performance of large target size and complex ground background in low-altitude UAV flight images

% 在此基础上，M4 模型在MSCA模块中的深度卷积移除 SiLU 激活函数，同时将 MSCA 模块保持在同一层（16）。此调整导致 mAP50 分数提升置 0.858，而不改变参数量。这与参考文献 [5] 的发现一致，在深度卷积中不建议加入非线性激活。

% 最后，M6 模型将 MSCA 模块移至第 22 层，同时仍然省略激活函数。此时mAP50下降至0.855,我们认为，这是由于第19层之后的特征图通过Proto NET编码的特征图更靠近最终输出，所以在18层加入MSCA 模块更有助于捕捉通道之间的多尺度特征信息

\subsubsection{Ablation experiments}
To evaluate the segmentation performance of the proposed ALSS-YOLO-Seg model, we conducted a series of ablation experiments, with results presented in Tables ~\ref{Ablationv1}, ~\ref{Ablationv2}. Each technical enhancement introduced into the model contributed to performance improvements. Beginning with the YOLOv8-Seg baseline (which retains the backbone and FPN of YOLOv8, while incorporating the same segmentation head as ALSS-YOLO-Seg.), we adjusted the model’s width hyperparameter to 0.18 in order to maintain a similar parameter count, and designated this variant as YOLOv8-Seg'. Although we also investigated adjustments to the depth hyperparameter, these yielded suboptimal results, with a \(mAP50_{mask}\) of 84.1\%.

\begin{figure}[h!]
\begin{center}
\includegraphics[width=17cm]{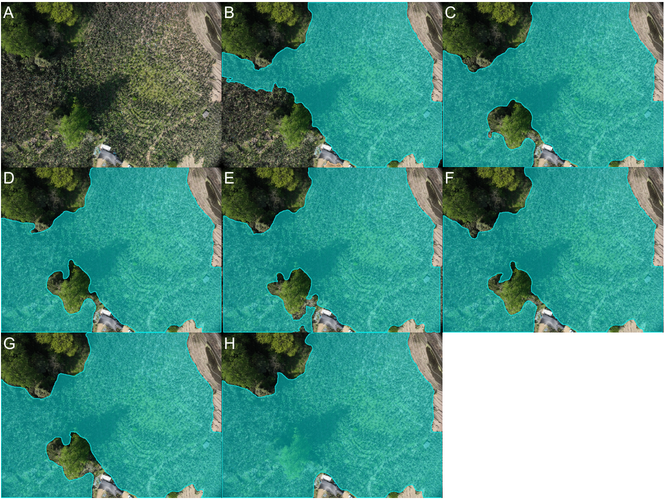}
\end{center}
\caption{Segmentation examples of different ablation experiments on the Banana Plantation dataset: (\textbf{a}) original image; (\textbf{b}) YOLOv8-Seg'; (\textbf{c}) M1; (\textbf{d}) M2; (\textbf{e}) M3; (\textbf{f}) M4; (\textbf{g}) M5(ALSS-YOLO-Seg) (\textbf{h}) M6.}\label{ablation_virsion}
\end{figure}

In the M1 model, the integration of the ALSS module improves the segmentation accuracy, with a \(mAP50_{mask}\) of 85.2\%. Importantly, the parameter count decreased to 1.5737 million, reflecting the dual benefit of performance enhancement and model efficiency. The ALSS module's contribution not only improved segmentation accuracy but also optimized parameter utilization, thereby reducing model complexity.

In the M2 model, a convolutional layer was added after the second layer of the network to enhance early-stage feature extraction. This adjustment resulted in a slight improvement in performance, with a \(mAP50_{mask}\) of 85.3\% and a parameter count of 1.5746 million. The added layer improved the model's ability to capture finer details without significantly increasing complexity.

Further refinement in the M3 model was achieved by adjusting the reg\_max value to 24, following the approach outlined in Ref. \citep{zhao2023fast}. This adjustment was particularly beneficial for improving the regression capability of bounding boxes, especially in handling the larger objects prevalent in the banana plantation dataset. As a result, the \(mAP50_{mask}\) increased to 85.5\%, with a corresponding rise in the parameter count to 1.8184 million. This demonstrates that fine-tuning specific hyperparameters can enhance segmentation performance, albeit with a moderate increase in model complexity.

In the exploration of the MSCA module's integration, the M3, M4, and M5 models were designed to investigate both its placement within the architecture and the effect of using or omitting activation functions on its performance.

The M3 model represents the initial stage, where the MSCA module was integrated at layer 16 with the use of the SiLU activation function. This modification resulted in a slight improvement in performance, with a \(mAP50_{mask}\) of 85.6\% and a parameter count of 1.8256 million. 

Building on this, the M4 model removed the SiLU activation function from the depthwise convolution in the MSCA module, while keeping the MSCA module at the same layer. This adjustment resulted in an improved \(mAP50_{mask}\) of 85.8\%, without increasing the parameter count. This observation is consistent with the findings of Ref \cite{Depthwise}, which suggest that non-linear activation functions may not be necessary in depthwise convolutions.

Lastly, in the M6 model, the MSCA module was moved to layer 22, while still omitting the activation function. This change, however, led to a decrease in the \(mAP50_{mask}\) to 85.5\%. We attribute this drop in performance to the fact that after layer 17, the feature maps encoded by ProtoNet are closer to the final output. Therefore, placing the MSCA module at layer 16 proves more effective for capturing multi-scale feature information across channels.

\begin{table*}[!h]
\centering
\caption{Ablation experiment results on Banana Plantation dataset. \label{Ablationv1}}
\resizebox{\textwidth}{!}{ % Resize the table to fit the text width
\begin{tabular}{cccccccccc}
\hline
\textbf{Model} & \textbf{ALSS} & \textbf{ADD} & \textbf{24} & \textbf{MSCA@L16} & \textbf{MSCA@L16} & \textbf{MSCA@L22} & \textbf{\( \text{mAP50}_{\text{mask}} \)} & \textbf{Params} & \textbf{FPS} \\
 & 
& \textbf{Conv} & \textbf{reg\_max} & \textbf{with SiLU} &  &  &  & \textbf{(M)} &  \\ 
\hline
YOLOv8-Seg' &  &  &  &  &  &  & 0.848 & 1,6952 & 126.2 \\ 
M1 & \checkmark &  &  &  &  &  & 0.852 & 1,5737 & 138.4 \\ 
M2 & \checkmark & \checkmark &  &  &  &  & 0.853 & 1,5746 & 147.8 \\ 
M3 & \checkmark & \checkmark & \checkmark &  &  &  & 0.855 & 1,8184 & 141.4 \\ 
M4 & \checkmark & \checkmark & \checkmark & \checkmark &  &  & 0.856 & 1,8256 & 133.8 \\ 
M5 (ALSS-YOLO-Seg) & \checkmark & \checkmark & \checkmark &  & \checkmark &  & 0.858 & 1,8256 & 130.6 \\ 
M6 & \checkmark & \checkmark & \checkmark &  &  & \checkmark & 0.855 & 1,8658 & 127.9 \\ 
\hline
\end{tabular}}
\noindent{\footnotesize{MSCA@L16 denotes the integration of the MSCA module at layer 16, while MSCA@L22 represents the integration of the MSCA module at layer 22.}}
\end{table*}

\begin{table*}[!h]
\centering
\caption{Extended performance comparison based on the Banana Plantation dataset. \label{Ablationv2}}
\resizebox{\textwidth}{!}{ % Resize the table to fit the text width
\begin{tabular}{cccccccccc}
\hline
\textbf{Model} & \textbf{\(P_{box}\)} & \textbf{\(R_{box}\)} & {mAP50} & {mAP50:90} & \textbf{\(P_{mask}\)} & \textbf{\(R_{mask}\)} & {mAP50} & {mAP50:90} \\
 &  &  & \textbf{\(_{box}\)} & \textbf{\(_{box}\)} &  &  & \textbf{\(_{mask}\)} & \textbf{\(_{mask}\)} \\

\hline
YOLOv8-Seg' & 0.897 & 0.823 & 0.846 & 0.751 & 0.910 & 0.825 & 0.848 & 0.721 \\ 
M1 & 0.893 & 0.827 & 0.844 & 0.736 & 0.899 & 0.833 & 0.852 & 0.725 \\ 
M2 & 0.919 & 0.815 & 0.854 & 0.738 & 0.914 & 0.828 & 0.853 & 0.732 \\ 
M3 & 0.917 & 0.817 & 0.856 & 0.742 & 0.912 & 0.818 & 0.855 & 0.726 \\ 
M4 & 0.917 & 0.818 & 0.857 & 0.742 & 0.919 & 0.821 & 0.856 & 0.731 \\ 
M5 (ALSS-YOLO-Seg) & 0.906 & 0.823 & 0.856 & 0.746 & 0.900 & 0.826 & 0.858 & 0.736 \\ 
M6 & 0.890 & 0.828 & 0.852 & 0.740 & 0.892 & 0.829 & 0.855 & 0.728 \\ 
\hline
\end{tabular}}
\end{table*}

To further validate the model in the ablation study, we selected a challenging scene with severe occlusion for segmentation evaluation on the test set, as shown in Figure ~\ref{ablation_virsion}. The segmentation results of all algorithms were obtained with a confidence threshold of 0.3. The results reveal that the YOLOv8-Seg' model performed the worst in this highly irregular and occluded scene, failing to segment the banana plantation areas effectively, leading to a low recall rate. Additionally, the M6 model struggled with accuracy, showing limited ability to differentiate tree regions. In contrast, the other models demonstrated more robust performance, effectively handling the occlusion and providing more accurate segmentation results.

\subsubsection{Comparison experiments}

\begin{figure}[h!]
\begin{center}
\includegraphics[width=17cm]{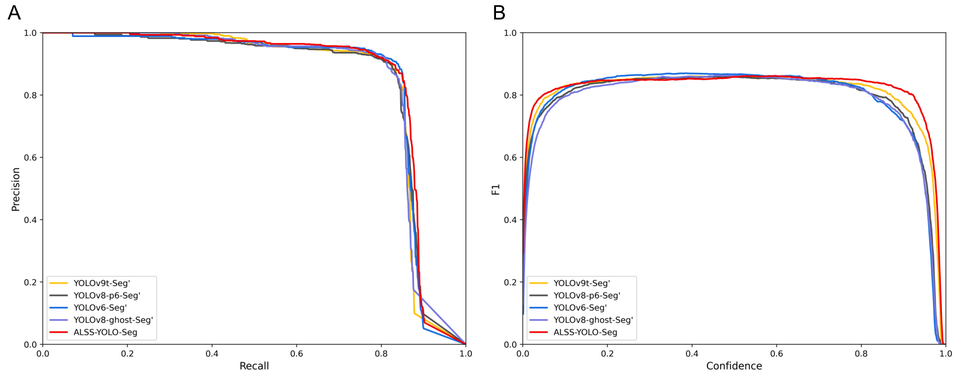}
\end{center}
\caption{Performance comparison chart of comparison experiment based on Banana Plantation dataset: (\textbf{a}) Precision-recall fitting curve of segmentation mask; (\textbf{b}) The F1 value of the segmentation mask varies with the confidence threshold.}
\label{comparison_curve}
\end{figure}

To validate the performance of the ALSS-YOLO-Seg model, we conducted a comprehensive comparison against other state-of-the-art segmentation models on the Banana Plantation dataset. These models include YOLOv3-tiny-Seg, YOLOv5-Seg', YOLOv6-Seg', YOLOv8-Seg', YOLOv8-ghost-Seg', YOLOv8-p2-Seg', YOLOv8-p6-Seg', YOLOv9t-Seg', and YOLOv10-Seg'. For a fair comparison, the prime (`) symbol indicates that the width hyperparameters of each model were adjusted to maintain a similar total number of parameters, ensuring that differences in performance are primarily due to variations in model architecture rather than model size. All models were trained under identical training environments and conditions. The comparison results of ALSS-YOLO-Seg with the other models are presented in Table~\ref{comparison-table}, while Figure~\ref{comparison_curve} visualizes the performance comparison between ALSS-YOLO-Seg and the top-performing models.

% redmon2018yolov3
\begin{table*}[!h]
\centering
\caption{Comparative experiments based on the Banana Plantation dataset. \label{comparison-table}}
\resizebox{\textwidth}{!}{
\begin{tabular}{ccccccccccc}
\hline
\textbf{Model} & \textbf{\(P_{box}\)} & \textbf{\(R_{box}\)} & {mAP50} & {mAP50:90} & \textbf{\(P_{mask}\)} & \textbf{\(R_{mask}\)} & {mAP50} & {mAP50:90} & {Params} & {FPS} \\
 &  &  & \textbf{\(_{box}\)} & \textbf{\(_{box}\)} &  &  & \textbf{\(_{mask}\)} & \textbf{\(_{mask}\)} & (M)  &  \\
 \hline
YOLOv3-tiny-Seg\citep{redmon2018yolov3} & 0.779 & 0.799 & 0.795 & 0.512 & 0.763 & 0.800 & 0.785 & 0.521 & 1.7975 & 270.6 \\
YOLOv5-Seg'      & 0.906 & 0.827 & 0.843 & 0.738 & 0.909 & 0.829 & 0.844 & 0.724 & 1.5196 & 119.3 \\
YOLOv6-Seg'\citep{li2022yolov6}    & 0.901 & 0.830 & 0.844 & 0.747 & 0.907 & 0.836 & 0.848 & 0.732 & 1.8963 & 148.5 \\
YOLOv8-Seg'\citep{yolov8_ultralytics} & 0.897 & 0.823 & 0.846 & 0.751 & 0.910 & 0.825 & 0.848 & 0.721 & 1.6952 & 126.2 \\ 
YOLOv8-ghost-Seg'\citep{yolov8_ultralytics} & 0.913 & 0.806 & 0.847 & 0.762 & 0.917 & 0.809 & 0.849 & 0.735 & 1.9206 & 138.1 \\
YOLOv8-p2-Seg'\citep{yolov8_ultralytics}   & 0.898 & 0.828 & 0.840 & 0.742 & 0.900 & 0.832 & 0.844 & 0.735 & 1.7447 & 101.6 \\
YOLOv8-p6-Seg'\citep{yolov8_ultralytics}   & 0.894 & 0.816 & 0.845 & 0.747 & 0.899 & 0.821 & 0.846 & 0.720 & 2.6950 & 102.5 \\
YOLOv9t-Seg'\citep{wang2024yolov9}    & 0.899 & 0.821 & 0.847 & 0.764 & 0.903 & 0.822 & 0.846 & 0.734 & 1.6619 & 82.6 \\
YOLOv10-Seg'\citep{wang2024yolov10}     & 0.894 & 0.824 & 0.843 & 0.741 & 0.893 & 0.823 & 0.838 & 0.717 & 1.5627 & 121.4 \\
ALSS-YOLO-Seg & 0.906 & 0.823 & 0.856 & 0.746 & 0.900 & 0.826 & 0.858 & 0.736 & 1.8256 & 130.6 \\ 
\hline
\end{tabular}}
\end{table*}

In the comparative experiments on the Banana Plantation dataset, the ALSS-YOLO-Seg model demonstrated strong segmentation performance, particularly in mask segmentation, achieving an impressive \(mAP50_{mask}\) of 85.8\%, as shown in Table~\ref{comparison-table}. This result is further confirmed by Figure~\ref{comparison_curve}A, where the red curve encloses the largest area. Moreover, Figure~\ref{comparison_curve}A and B clearly illustrate that ALSS-YOLO-Seg maintains a distinct advantage across both high and low confidence thresholds. This performance surpasses most other state-of-the-art segmentation models in the study. For instance, YOLOv3-tiny-Seg, despite being a lightweight model, achieved a considerably lower \(mAP50_{mask}\) of 78.5\%, highlighting the superior segmentation capabilities of ALSS-YOLO-Seg, particularly in the complex scenarios encountered in banana plantation environments.

Models such as YOLOv5-Seg', YOLOv6-Seg', YOLOv8-Seg', and YOLOv8-ghost-Seg' exhibited commendable performance, achieving \(mAP50_{mask}\) values of 84.4\%, 84.8\%, 84.8\%, and 84.9\%, respectively. ALSS-YOLO-Seg consistently outperformed these models, attaining a superior \(mAP50_{mask}\) of 85.8\% while maintaining a competitive parameter count. Additionally, YOLOv8-p6-Seg and YOLOv9t-Seg demonstrated strong performance with \(mAP50_{mask}\) values of 84.6\% each; however, they did not match the performance of ALSS-YOLO-Seg. Furthermore, YOLOv10-Seg' achieved an \(mAP50_{mask}\) of 83.8\%, which also falls short compared to ALSS-YOLO-Seg. Given the already high \(mAP50_{mask}\) across the evaluated models, the notable enhancement provided by ALSS-YOLO-Seg is particularly significant.

ALSS-YOLO-Seg’s high \(mAP50_{mask}\) highlights its robustness and effectiveness in handling complex segmentation tasks such as occlusions and complex backgrounds in the banana plantation dataset. The superior mask accuracy makes ALSS-YOLO-Seg particularly suitable for real-world applications that require accurate segmentation, especially in UAV-based agricultural monitoring scenarios where accuracy and computational efficiency are critical.

\begin{figure}[h!]
\begin{center}
\includegraphics[width=17cm]{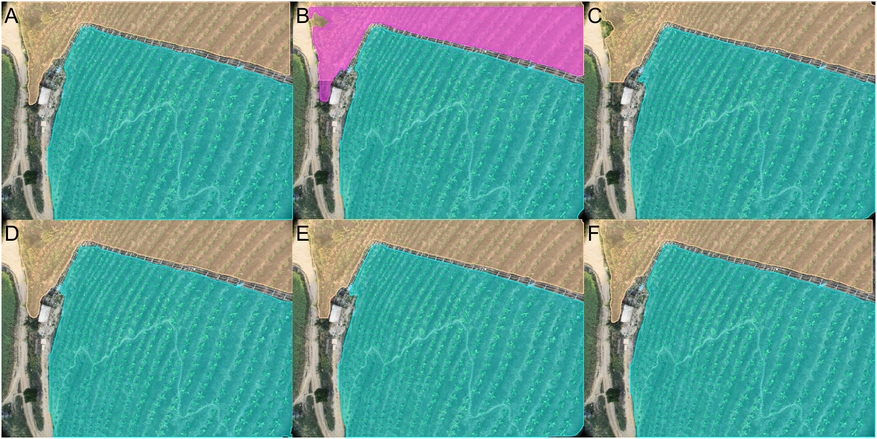}
\end{center}
\caption{Segmentation performance under occlusion by banana transport cable way: (\textbf{a}) ground truth; (\textbf{b}) YOLOv9t-Seg'; (\textbf{c}) YOLOv8-p6-Seg'; (\textbf{d}) YOLOv6-Seg'; (\textbf{e}) YOLOv8-ghost-Seg'; (\textbf{f}) ALSS-YOLO-Seg.}\label{occlusion}
\end{figure}

\begin{figure}[h!]
\begin{center}
\includegraphics[width=17cm]{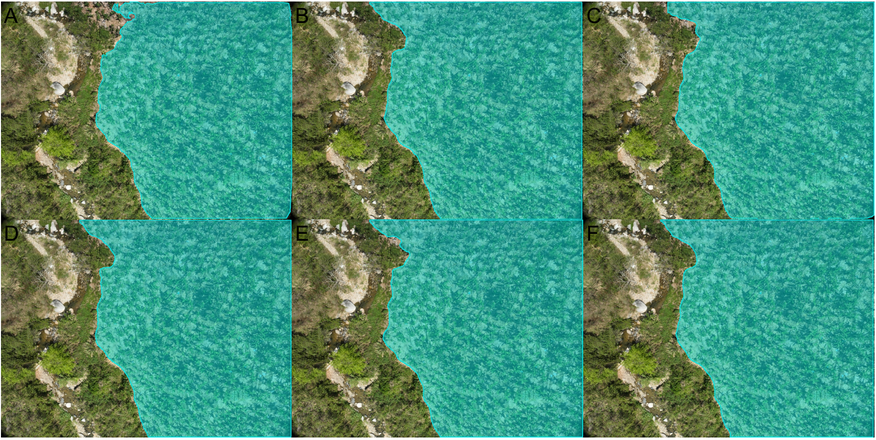}
\end{center}
\caption{Segmentation performance with weed interference: (\textbf{a}) ground truth; (\textbf{b}) YOLOv9t-Seg'; (\textbf{c}) YOLOv8-p6-Seg'; (\textbf{d}) YOLOv6-Seg'; (\textbf{e}) YOLOv8-ghost-Seg'; (\textbf{f}) ALSS-YOLO-Seg.}\label{weeds}
\end{figure}

\begin{figure}[h!]
\begin{center}
\includegraphics[width=17cm]{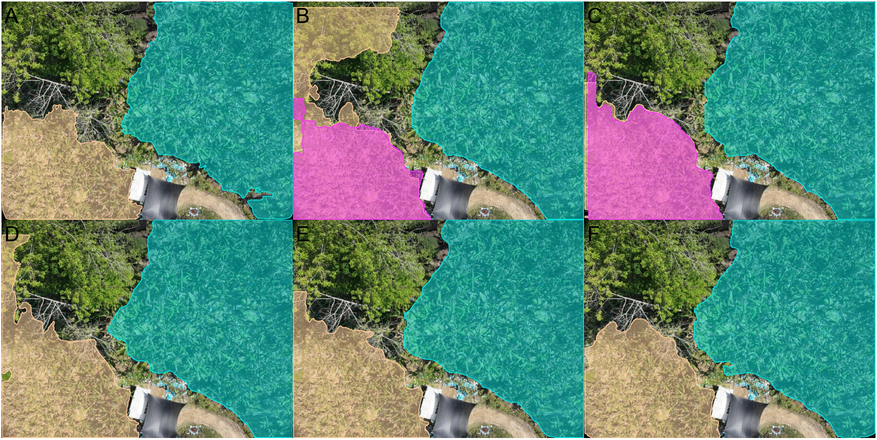}
\end{center}
\caption{Segmentation performance of large target size and complex ground background in low-altitude UAV flight images: (\textbf{a}) ground truth; (\textbf{b}) YOLOv9t-Seg'; (\textbf{c}) YOLOv8-p6-Seg'; (\textbf{d}) YOLOv6-Seg'; (\textbf{e}) YOLOv8-ghost-Seg'; (\textbf{f}) ALSS-YOLO-Seg.}\label{sizes}
\end{figure}

In Figures ~\ref{occlusion}, ~\ref{weeds}, and ~\ref{sizes}, we selected G, H, and I from Figure~\ref{banana} to qualitatively evaluate the segmentation performance of various advanced models in complex scenarios.

In Figure~\ref{occlusion}, it is evident that the segmentation results produced by ALSS-YOLO-Seg exhibit the highest accuracy, closely aligning with the ground truth. In contrast, the YOLOv9t-Seg' model exhibits significant errors, resulting in overlapping segmentation masks within the same region. Figure~\ref{weeds} illustrates that all models demonstrate a commendable ability to distinguish between banana plantations and weed areas. Notably, ALSS-YOLO-Seg achieves a competitive segmentation performance, even surpassing the segmentation masks generated by the iterative optimization annotation pipeline. 
The superior performance of ALSS-YOLO-Seg can be attributed to the extensive use of the Banana Plantation dataset during training, which enables the model to effectively capture the semantic features specific to this environment. By training on a comprehensive set of annotated data, ALSS-YOLO-Seg is capable of identifying subtle differences between banana plants and weeds, thus mitigating the risk of overfitting to individual images. 
Furthermore, while we observed that tuning down the parameter \(\epsilon\) mentioned in Section 2.2 can improve the quality of the segmentation masks generated by the iterative optimization annotation pipeline, but this tuning requires additional iterations to achieve better results. Importantly, the quality of our banana plantation dataset is already high enough compared to traditional manually annotated datasets such as COCO \citep{lin2014microsoft}\citep{kirillov2023segment}.
In Figure~\ref{sizes}, the segmentation performance of ALSS-YOLO-Seg is markedly superior to that of other models. For example, both YOLOv9t-Seg' and YOLOv8-p6-Seg' exhibit overlapping segmentation masks within the same region, indicating a failure to accurately distinguish between adjacent targets. Additionally, YOLOv6-Seg' struggles to differentiate between trees and banana plantations, incorrectly classifying portions of the tree regions as part of the banana plantations. YOLOv8-ghost-Seg' demonstrates a similar issue. These results suggest that ALSS-YOLO-Seg outperforms other models in this context due to the integration of the ALSS module in its backbone, which enhances inter-channel information exchange and optimizes feature representation. Furthermore, the inclusion of the MSCA effectively supports multi-scale feature extraction and attention refinement, thereby addressing challenges related to varying target sizes and complex backgrounds.

In Table~\ref{comparison-tablev2}, we present additional comparative experiments to comprehensively evaluate the performance of various segmentation models on the Banana Plantation dataset. The table lists key metrics such as segmentation accuracy (mIoU) and parameter count for seven different models.

In terms of segmentation accuracy (mIoU), KNet (Swin-L) achieves the best performance with an mIoU of 94.68\%, indicating its strong ability to capture complex features in banana plantation scenes. Following closely is Mask2Former (Swin-L) with an mIoU of 94.27\%. Although slightly lower than KNet, it still demonstrates robust segmentation capabilities. However, these two models have parameter counts of 245M and 235M, respectively, indicating their high computational demands, which may limit their use in real-time applications or environments with limited computational power, such as UAV-based platforms.

In contrast, our proposed ALSS-YOLO-Seg model achieves an impressive mIoU of 92.72\%, slightly lower than KNet and Mask2Former, but with a marginal performance gap. More importantly, ALSS-YOLO-Seg has a significantly lower parameter count of only 1.83M, far below the other models. This demonstrates an excellent balance between accuracy and model complexity. The low parameter count makes ALSS-YOLO-Seg highly efficient for deployment on platforms with limited computational resources.

Additionally, traditional segmentation models like DeepLabV3+ (R-101-D8) also achieve commendable segmentation accuracy, with an mIoU of 92.83\%. However, its parameter count reaches 60.21M, making it considerably more complex than ALSS-YOLO-Seg and less suitable for deployment on lightweight hardware. Similarly, PSPNet (R-50-D8) and UNet (UNet-S5-D16) achieve mIoU of 91.13\% and 90.31\%, respectively. While their parameter counts are somewhat lower (46.60M and 28.99M, respectively), their segmentation performance may still present challenges in meeting the high precision demands required in banana plantation monitoring.

In summary, our proposed ALSS-YOLO-Seg model exhibits excellent segmentation accuracy, especially in UAV-based banana plantation scenarios, while significantly reducing the number of parameters. The model achieves an excellent balance between accuracy and computational efficiency, making it ideal for agricultural monitoring tasks with stringent requirements for real-time performance and resource efficiency on UAV-based platforms or other low-power edge devices.

\begin{table*}[!h]
\centering
\caption{Performance comparison based on the Banana Plantation dataset.\label{comparison-tablev2}}
% \resizebox{\textwidth}{!}{
\begin{tabular}{ccccc}
\hline
\textbf{Model} & \textbf{Backbone} & \textbf{mIoU} & \textbf{Params (M)} & \textbf{FPS} \\
\hline
DeepLabV3plus\citep{chen2018encoder} & R-101-D8 & 0.92.83 & 60.21 & 6.98 \\
KNet\citep{zhang2021k} & Swin-L & 0.9468 & 245 & 4.55 \\
Mask2Former\citep{cheng2022masked} & Swin-L & 0.9427 & 235 & 5.09 \\
PSPNet\citep{zhao2017pyramid} & R-50-D8 & 0.9113 & 46.60 & 11.91 \\
Segformer\citep{xie2021segformer} & MIT-B5 & 0.9398 & 81.97 & 6.57 \\
UNet\citep{ronneberger2015u} & UNet-S5-D16 & 0.9031 & 28.99 & 17.93 \\
ALSS-YOLO-Seg & ALSSNet & 0.9272 & 1.8256 & 130.6 \\
\hline
\end{tabular}
\end{table*}

\section{Conclusion}

In this research, we proposed a comprehensive solution for UAV-based segmentation of banana plantations by integrating an efficient annotation pipeline with a lightweight segmentation model. First, we developed an iterative optimization annotation pipeline that significantly reduces data annotation costs and time, demonstrating the ability to generate high-quality segmentation masks from a limited number of weak annotations. This approach minimizes manual intervention while enhancing annotation efficiency and segmentation accuracy. Our method achieved high-quality mask generation on the Javeri dataset, with an \(mIoU\) of 93.8\%. This pipeline was also utilized to construct a custom banana plantation dataset.
Subsequently, we trained a specialized lightweight and efficient segmentation model, ALSS-YOLO-Seg, using this dataset. The model incorporates the Adaptive Lightweight Channel Splitting and Shuffling (ALSS) module to enhance information exchange between channels and optimize feature extraction, thus aiding in accurate crop identification. The architecture employs a bottleneck structure to reduce complexity and ensure efficient processing. Additionally, the Multi-Scale Channel Attention (MSCA) module combines multi-scale feature extraction with channel attention mechanisms, significantly improving the model's ability to handle varying target sizes and complex backgrounds by focusing on the most relevant features across different scales. 

Extensive experiments conducted on our custom banana plantation dataset demonstrated that ALSS-YOLO-Seg achieved state-of-the-art performance, with an \(mAP50_{mask}\) of 85.8\% and a parameter count of 1.8256M, surpassing the baseline YOLOv8-Seg model, which attained an \(mAP50_{mask}\) of 84.8\%. The proposed approach effectively addresses the challenges associated with large-scale data annotation and ensures high segmentation accuracy in resource-constrained environments, making it highly suitable for real-world agricultural applications, such as crop monitoring and plantation management.

For future work, we aim to further enhance the generalization capabilities of the ALSS-YOLO-Seg model by incorporating additional contextual information and exploring more advanced attention mechanisms. Expanding the custom banana plantation dataset with diverse environmental conditions and varying stages of crop growth will allow for more robust model training and improved performance in different settings. Additionally, the transferability of this approach to other agricultural sectors will be explored to broaden its applicability across a range of crops and plantation types.

% \paragraph{Level 4}
% \subparagraph{Level 5}

\section*{Data availability statement}
The datasets utilized and analyzed during the current study are available without restriction and can be accessed at the following link: https://github.com/helloworlder8/computer\_vision.

\section*{Author contributions}
% YW: Conceptualization, Data curation, Formal analysis, Funding acquisition, Investigation, Methodology, Project administration, Resources, Software, Supervision, Validation, Visualization, Writing – original draft, Writing – review & editing. 
% ZL: Formal analysis, Investigation, Resources, Software, Supervision, Validation, Writing – original draft, Writing – review & editing. 
% HJ: Data curation, Investigation, Software, Visualization, Writing – review & editing. QL: Methodology, Project administration, Supervision, Writing – review & editing. 
% JQ: Investigation, Resources, Writing review & editing. 
% FP: Resources, Visualization, Writing – review & editing. 
% XF: Formal analysis, Project administration, Supervision, Writing – review & editing. 
% BG: Funding acquisition, Project administration, Supervision, Writing – review & editing.
Ang He: Conceptualization, Data curation, Formal analysis, Investigation, Methodology, Project administration, Resources, Software, Validation, Visualization, Writing - original draft, Writing - review \& editing. 
Ximei Wu: Conceptualization, Formal analysis, Investigation, Validation. 
Xing Xu: Conceptualization, Investigation, Data curation, Writing - review \& editing. 
Jing Chen: Conceptualization, Resources, Writing - review \& editing. 
Xiaobin Guo: Conceptualization, Funding acquisition, Project administration, Supervision, Writing - review \& editing. 
Sheng Xu: Conceptualization, Formal analysis, Project administration, Supervision, Writing - review \& editing. 

% xuxing@scau.edu.cn
% jchen125@gdut.edu.cn
% guoxb@gdut.edu.cn
% xusheng@gdut.edu.cn
\section*{Funding}
This work was supported by the National Natural Science Foundation of China (No. 11904056). Guangzhou Basic and Applied Basic Research Project (No. 202102020053).

\section*{Conflict of interest}
The authors declare that the research was conducted in the absence of any commercial or financial relationships that could be construed as a potential conflict of interest.

\section*{Publisher’s note}
All claims expressed in this article are solely those of the authors and do not necessarily represent those of their affiliated organizations, or those of the publisher, the editors and the reviewers. Any product that may be evaluated in this article, or claim that may be made by its manufacturer, is not guaranteed or endorsed by the publisher.

% \section*{Supplemental Data}
%  \href{http://home.frontiersin.org/about/author-guidelines#SupplementaryMaterial}{Supplementary Material} should be uploaded separately on submission, if there are Supplementary Figures, please include the caption in the same file as the figure. LaTeX Supplementary Material templates can be found in the Frontiers LaTeX folder.

% \section*{Data Availability Statement}
% The datasets [GENERATED/ANALYZED] for this study can be found in the [NAME OF REPOSITORY] [LINK].
% Please see the availability of data guidelines for more information, at https://www.frontiersin.org/about/author-guidelines#AvailabilityofData

\bibliographystyle{Frontiers-Harvard} %  Many Frontiers journals use the Harvard referencing system (Author-date), to find the style and resources for the journal you are submitting to: https://zendesk.frontiersin.org/hc/en-us/articles/360017860337-Frontiers-Reference-Styles-by-Journal. For Humanities and Social Sciences articles please include page numbers in the in-text citations 
\bibliography{test}

\begin{thebibliography}{62}
\providecommand{\natexlab}[1]{#1}
\expandafter\ifx\csname urlstyle\endcsname\relax
  \providecommand{\doi}[1]{doi:\discretionary{}{}{}#1}\else
  \providecommand{\doi}{doi:\discretionary{}{}{}\begingroup
  \urlstyle{rm}\Url}\fi
\providecommand{\selectlanguage}[1]{\relax}
\providecommand{\bibAnnoteFile}[1]{%
  \IfFileExists{#1}{\begin{quotation}\noindent\textsc{Key:} #1\\
  \textsc{Annotation:}\ \input{#1}\end{quotation}}{}}
\providecommand{\bibAnnote}[2]{%
  \begin{quotation}\noindent\textsc{Key:} #1\\
  \textsc{Annotation:}\ #2\end{quotation}}

\bibitem[{Alexey(2020)}]{alexey2020image}
Alexey, D. (2020).
\newblock An image is worth 16x16 words: Transformers for image recognition at
  scale.
\newblock \emph{arXiv preprint arXiv: 2010.11929}
\bibAnnoteFile{alexey2020image}

\bibitem[{Bolya et~al.(2019)Bolya, Zhou, Xiao, and Lee}]{bolya2019yolact}
Bolya, D., Zhou, C., Xiao, F., and Lee, Y.~J. (2019).
\newblock Yolact: Real-time instance segmentation.
\newblock In \emph{Proceedings of the IEEE/CVF international conference on
  computer vision}. 9157--9166
\bibAnnoteFile{bolya2019yolact}

\bibitem[{Cai and Vasconcelos(2019)}]{cai2019cascade}
Cai, Z. and Vasconcelos, N. (2019).
\newblock Cascade r-cnn: High quality object detection and instance
  segmentation.
\newblock \emph{IEEE transactions on pattern analysis and machine intelligence}
  43, 1483--1498
\bibAnnoteFile{cai2019cascade}

\bibitem[{Chen et~al.(2024)Chen, Hou, Cui, Li, Shangguan, and
  Cao}]{chen2024yolov8}
Chen, G., Hou, Y., Cui, T., Li, H., Shangguan, F., and Cao, L. (2024).
\newblock Yolov8-cml: A lightweight target detection method for color-changing
  melon ripening in intelligent agriculture.
\newblock \emph{Scientific Reports} 14, 14400
\bibAnnoteFile{chen2024yolov8}

\bibitem[{Chen et~al.(2020)Chen, Sun, Tian, Shen, Huang, and
  Yan}]{chen2020blendmask}
Chen, H., Sun, K., Tian, Z., Shen, C., Huang, Y., and Yan, Y. (2020).
\newblock Blendmask: Top-down meets bottom-up for instance segmentation.
\newblock In \emph{Proceedings of the IEEE/CVF conference on computer vision
  and pattern recognition}. 8573--8581
\bibAnnoteFile{chen2020blendmask}

\bibitem[{Chen et~al.(2018)Chen, Zhu, Papandreou, Schroff, and
  Adam}]{chen2018encoder}
Chen, L.-C., Zhu, Y., Papandreou, G., Schroff, F., and Adam, H. (2018).
\newblock Encoder-decoder with atrous separable convolution for semantic image
  segmentation.
\newblock In \emph{Proceedings of the European conference on computer vision
  (ECCV)}. 801--818
\bibAnnoteFile{chen2018encoder}

\bibitem[{Chen et~al.(2022)Chen, Duan, Wang, He, Lu, Dai
  et~al.}]{chen2022vision}
Chen, Z., Duan, Y., Wang, W., He, J., Lu, T., Dai, J., et~al. (2022).
\newblock Vision transformer adapter for dense predictions.
\newblock \emph{arXiv preprint arXiv:2205.08534}
\bibAnnoteFile{chen2022vision}

\bibitem[{Cheng et~al.(2022)Cheng, Misra, Schwing, Kirillov, and
  Girdhar}]{cheng2022masked}
Cheng, B., Misra, I., Schwing, A.~G., Kirillov, A., and Girdhar, R. (2022).
\newblock Masked-attention mask transformer for universal image segmentation.
\newblock In \emph{Proceedings of the IEEE/CVF conference on computer vision
  and pattern recognition}. 1290--1299
\bibAnnoteFile{cheng2022masked}

\bibitem[{Chollet(2017)}]{Depthwise}
Chollet, F. (2017).
\newblock Xception: Deep learning with depthwise separable convolutions.
\newblock In \emph{Proceedings of the IEEE conference on computer vision and
  pattern recognition}. 1251--1258
\bibAnnoteFile{Depthwise}

\bibitem[{De~Barros et~al.(2009)De~Barros, Blazy, Rodrigues, Tournebize, and
  Cinna}]{de2009emergy}
De~Barros, I., Blazy, J.~M., Rodrigues, G.~S., Tournebize, R., and Cinna, J.~P.
  (2009).
\newblock Emergy evaluation and economic performance of banana cropping systems
  in guadeloupe (french west indies).
\newblock \emph{Agriculture, Ecosystems \& Environment} 129, 437--449
\bibAnnoteFile{de2009emergy}

\bibitem[{de~Souza et~al.(2019)de~Souza, Neto, Piazentin, Junior, Gomes, Bonini
  et~al.}]{de2019artificial}
de~Souza, A.~V., Neto, A.~B., Piazentin, J.~C., Junior, B. J.~D., Gomes, E.~P.,
  Bonini, C. d. S.~B., et~al. (2019).
\newblock Artificial neural network modelling in the prediction of bananas’
  harvest.
\newblock \emph{Scientia Horticulturae} 257, 108724
\bibAnnoteFile{de2019artificial}

\bibitem[{D{\'\i}az et~al.(2011)D{\'\i}az, P{\'e}rez, Mateos, Marinescu, and
  Guerra}]{diaz2011novel}
D{\'\i}az, S.~E., P{\'e}rez, J.~C., Mateos, A.~C., Marinescu, M.-C., and
  Guerra, B.~B. (2011).
\newblock A novel methodology for the monitoring of the agricultural production
  process based on wireless sensor networks.
\newblock \emph{Computers and electronics in agriculture} 76, 252--265
\bibAnnoteFile{diaz2011novel}

\bibitem[{He et~al.(2024)He, Li, Wu, Su, Chen, Xu et~al.}]{10680397}
He, A., Li, X., Wu, X., Su, C., Chen, J., Xu, S., et~al. (2024).
\newblock Alss-yolo: An adaptive lightweight channel split and shuffling
  network for tir wildlife detection in uav imagery.
\newblock \emph{IEEE Journal of Selected Topics in Applied Earth Observations
  and Remote Sensing} 17, 17308--17326.
\newblock \doi{10.1109/JSTARS.2024.3461172}
\bibAnnoteFile{10680397}

\bibitem[{He et~al.(2017)He, Gkioxari, Doll{\'a}r, and Girshick}]{he2017mask}
He, K., Gkioxari, G., Doll{\'a}r, P., and Girshick, R. (2017).
\newblock Mask r-cnn.
\newblock In \emph{Proceedings of the IEEE international conference on computer
  vision}. 2961--2969
\bibAnnoteFile{he2017mask}

\bibitem[{Hu et~al.(2018)Hu, Shen, and Sun}]{hu2018squeeze}
Hu, J., Shen, L., and Sun, G. (2018).
\newblock Squeeze-and-excitation networks.
\newblock In \emph{Proceedings of the IEEE conference on computer vision and
  pattern recognition}. 7132--7141
\bibAnnoteFile{hu2018squeeze}

\bibitem[{Huang et~al.(2019)Huang, Huang, Gong, Huang, and
  Wang}]{huang2019mask}
Huang, Z., Huang, L., Gong, Y., Huang, C., and Wang, X. (2019).
\newblock Mask scoring r-cnn.
\newblock In \emph{Proceedings of the IEEE/CVF conference on computer vision
  and pattern recognition}. 6409--6418
\bibAnnoteFile{huang2019mask}

\bibitem[{Ioffe(2015)}]{ioffe2015batch}
Ioffe, S. (2015).
\newblock Batch normalization: Accelerating deep network training by reducing
  internal covariate shift.
\newblock \emph{arXiv preprint arXiv:1502.03167}
\bibAnnoteFile{ioffe2015batch}

\bibitem[{Jain et~al.(2023)Jain, Singh, Orlov, Huang, Li, Walton
  et~al.}]{jain2023semask}
Jain, J., Singh, A., Orlov, N., Huang, Z., Li, J., Walton, S., et~al. (2023).
\newblock Semask: Semantically masked transformers for semantic segmentation.
\newblock In \emph{Proceedings of the IEEE/CVF International Conference on
  Computer Vision}. 752--761
\bibAnnoteFile{jain2023semask}

\bibitem[{Jia et~al.(2021)Jia, Yang, Xia, Chen, Parekh, Pham
  et~al.}]{jia2021scaling}
Jia, C., Yang, Y., Xia, Y., Chen, Y.-T., Parekh, Z., Pham, H., et~al. (2021).
\newblock Scaling up visual and vision-language representation learning with
  noisy text supervision.
\newblock In \emph{International conference on machine learning} (PMLR),
  4904--4916
\bibAnnoteFile{jia2021scaling}

\bibitem[{Jocher et~al.(2023)Jocher, Chaurasia, and Qiu}]{yolov8_ultralytics}
[Dataset] Jocher, G., Chaurasia, A., and Qiu, J. (2023).
\newblock Ultralytics yolov8
\bibAnnoteFile{yolov8_ultralytics}

\bibitem[{Kirillov et~al.(2023)Kirillov, Mintun, Ravi, Mao, Rolland, Gustafson
  et~al.}]{kirillov2023segment}
Kirillov, A., Mintun, E., Ravi, N., Mao, H., Rolland, C., Gustafson, L., et~al.
  (2023).
\newblock Segment anything.
\newblock In \emph{Proceedings of the IEEE/CVF International Conference on
  Computer Vision}. 4015--4026
\bibAnnoteFile{kirillov2023segment}

\bibitem[{Li et~al.(2022{\natexlab{a}})Li, Li, Jiang, Weng, Geng, Li
  et~al.}]{li2022yolov6}
Li, C., Li, L., Jiang, H., Weng, K., Geng, Y., Li, L., et~al.
  (2022{\natexlab{a}}).
\newblock Yolov6: A single-stage object detection framework for industrial
  applications.
\newblock \emph{arXiv preprint arXiv:2209.02976}
\bibAnnoteFile{li2022yolov6}

\bibitem[{Li et~al.(2023{\natexlab{a}})Li, Zhang, Xu, Liu, Zhang, Ni
  et~al.}]{li2023mask}
Li, F., Zhang, H., Xu, H., Liu, S., Zhang, L., Ni, L.~M., et~al.
  (2023{\natexlab{a}}).
\newblock Mask dino: Towards a unified transformer-based framework for object
  detection and segmentation.
\newblock In \emph{Proceedings of the IEEE/CVF Conference on Computer Vision
  and Pattern Recognition}. 3041--3050
\bibAnnoteFile{li2023mask}

\bibitem[{Li et~al.(2022{\natexlab{b}})Li, Ji, Qu, Zhou, and Cao}]{li2022cross}
Li, G., Ji, Z., Qu, X., Zhou, R., and Cao, D. (2022{\natexlab{b}}).
\newblock Cross-domain object detection for autonomous driving: A stepwise
  domain adaptative yolo approach.
\newblock \emph{IEEE Transactions on Intelligent Vehicles} 7, 603--615
\bibAnnoteFile{li2022cross}

\bibitem[{Li et~al.(2023{\natexlab{b}})Li, Zheng, Li, Long, Li, and
  Gao}]{li2023tomato}
Li, P., Zheng, J., Li, P., Long, H., Li, M., and Gao, L. (2023{\natexlab{b}}).
\newblock Tomato maturity detection and counting model based on mhsa-yolov8.
\newblock \emph{Sensors} 23, 6701
\bibAnnoteFile{li2023tomato}

\bibitem[{Li et~al.(2023{\natexlab{c}})Li, Xu, Wu, Zhao, Chen, Lu
  et~al.}]{li2023yolo}
Li, Z., Xu, B., Wu, D., Zhao, K., Chen, S., Lu, M., et~al.
  (2023{\natexlab{c}}).
\newblock A yolo-ggcnn based grasping framework for mobile robots in unknown
  environments.
\newblock \emph{Expert Systems with Applications} 225, 119993
\bibAnnoteFile{li2023yolo}

\bibitem[{Lin et~al.(2017)Lin, Doll{\'a}r, Girshick, He, Hariharan, and
  Belongie}]{lin2017feature}
Lin, T.-Y., Doll{\'a}r, P., Girshick, R., He, K., Hariharan, B., and Belongie,
  S. (2017).
\newblock Feature pyramid networks for object detection.
\newblock In \emph{Proceedings of the IEEE conference on computer vision and
  pattern recognition}. 2117--2125
\bibAnnoteFile{lin2017feature}

\bibitem[{Lin et~al.(2014)Lin, Maire, Belongie, Hays, Perona, Ramanan
  et~al.}]{lin2014microsoft}
Lin, T.-Y., Maire, M., Belongie, S., Hays, J., Perona, P., Ramanan, D., et~al.
  (2014).
\newblock Microsoft coco: Common objects in context.
\newblock In \emph{Computer Vision--ECCV 2014: 13th European Conference,
  Zurich, Switzerland, September 6-12, 2014, Proceedings, Part V 13}
  (Springer), 740--755
\bibAnnoteFile{lin2014microsoft}

\bibitem[{Liu et~al.(2022)Liu, Hu, Lin, Yao, Xie, Wei et~al.}]{liu2022swin}
Liu, Z., Hu, H., Lin, Y., Yao, Z., Xie, Z., Wei, Y., et~al. (2022).
\newblock Swin transformer v2: Scaling up capacity and resolution.
\newblock In \emph{Proceedings of the IEEE/CVF conference on computer vision
  and pattern recognition}. 12009--12019
\bibAnnoteFile{liu2022swin}

\bibitem[{Lund and Wang(2023)}]{lund2023chatting}
Lund, B.~D. and Wang, T. (2023).
\newblock Chatting about chatgpt: how may ai and gpt impact academia and
  libraries?
\newblock \emph{Library hi tech news} 40, 26--29
\bibAnnoteFile{lund2023chatting}

\bibitem[{Mo et~al.(2021)Mo, Lan, Yang, Wen, Qiu, Chen et~al.}]{mo2021deep}
Mo, J., Lan, Y., Yang, D., Wen, F., Qiu, H., Chen, X., et~al. (2021).
\newblock Deep learning-based instance segmentation method of litchi canopy
  from uav-acquired images.
\newblock \emph{Remote Sensing} 13, 3919
\bibAnnoteFile{mo2021deep}

\bibitem[{Nguyen et~al.(2021)Nguyen, Ta, Nguyen, Pham, Nguyen
  et~al.}]{nguyen2021yolo}
Nguyen, H.~H., Ta, T.~N., Nguyen, N.~C., Pham, H.~M., Nguyen, D.~M., et~al.
  (2021).
\newblock Yolo based real-time human detection for smart video surveillance at
  the edge.
\newblock In \emph{2020 IEEE eighth international conference on communications
  and electronics (ICCE)} (IEEE), 439--444
\bibAnnoteFile{nguyen2021yolo}

\bibitem[{Qin et~al.(2023)Qin, Zhang, Zhang, Chen, Yasunaga, and
  Yang}]{qin2023chatgpt}
Qin, C., Zhang, A., Zhang, Z., Chen, J., Yasunaga, M., and Yang, D. (2023).
\newblock Is chatgpt a general-purpose natural language processing task solver?
\newblock \emph{arXiv preprint arXiv:2302.06476}
\bibAnnoteFile{qin2023chatgpt}

\bibitem[{Radford et~al.(2021)Radford, Kim, Hallacy, Ramesh, Goh, Agarwal
  et~al.}]{radford2021learning}
Radford, A., Kim, J.~W., Hallacy, C., Ramesh, A., Goh, G., Agarwal, S., et~al.
  (2021).
\newblock Learning transferable visual models from natural language
  supervision.
\newblock In \emph{International conference on machine learning} (PMLR),
  8748--8763
\bibAnnoteFile{radford2021learning}

\bibitem[{Ramesh et~al.(2021)Ramesh, Pavlov, Goh, Gray, Voss, Radford
  et~al.}]{ramesh2021zero}
Ramesh, A., Pavlov, M., Goh, G., Gray, S., Voss, C., Radford, A., et~al.
  (2021).
\newblock Zero-shot text-to-image generation.
\newblock In \emph{International conference on machine learning} (Pmlr),
  8821--8831
\bibAnnoteFile{ramesh2021zero}

\bibitem[{Rao et~al.(2022)Rao, Zhao, Tang, Zhou, Lim, and Lu}]{rao2022hornet}
Rao, Y., Zhao, W., Tang, Y., Zhou, J., Lim, S.~N., and Lu, J. (2022).
\newblock Hornet: Efficient high-order spatial interactions with recursive
  gated convolutions.
\newblock \emph{Advances in Neural Information Processing Systems} 35,
  10353--10366
\bibAnnoteFile{rao2022hornet}

\bibitem[{Ravi et~al.(2024{\natexlab{a}})Ravi, Gabeur, Hu, Hu, Ryali, Ma
  et~al.}]{ravi2024sam}
Ravi, N., Gabeur, V., Hu, Y.-T., Hu, R., Ryali, C., Ma, T., et~al.
  (2024{\natexlab{a}}).
\newblock Sam 2: Segment anything in images and videos.
\newblock \emph{arXiv preprint arXiv:2408.00714}
\bibAnnoteFile{ravi2024sam}

\bibitem[{Ravi et~al.(2024{\natexlab{b}})Ravi, Gabeur, Hu, Hu, Ryali, Ma
  et~al.}]{ravi2024sam2}
Ravi, N., Gabeur, V., Hu, Y.-T., Hu, R., Ryali, C., Ma, T., et~al.
  (2024{\natexlab{b}}).
\newblock Sam 2: Segment anything in images and videos.
\newblock \emph{arXiv preprint arXiv:2408.00714}
\bibAnnoteFile{ravi2024sam2}

\bibitem[{Redmon(2018)}]{redmon2018yolov3}
Redmon, J. (2018).
\newblock Yolov3: An incremental improvement.
\newblock \emph{arXiv preprint arXiv:1804.02767}
\bibAnnoteFile{redmon2018yolov3}

\bibitem[{Ronneberger et~al.(2015)Ronneberger, Fischer, and
  Brox}]{ronneberger2015u}
Ronneberger, O., Fischer, P., and Brox, T. (2015).
\newblock U-net: Convolutional networks for biomedical image segmentation.
\newblock In \emph{Medical image computing and computer-assisted
  intervention--MICCAI 2015: 18th international conference, Munich, Germany,
  October 5-9, 2015, proceedings, part III 18} (Springer), 234--241
\bibAnnoteFile{ronneberger2015u}

\bibitem[{Sampurno et~al.(2024)Sampurno, Liu, Abeyrathna, and
  Ahamed}]{sampurno2024intrarow}
Sampurno, R.~M., Liu, Z., Abeyrathna, R. R.~D., and Ahamed, T. (2024).
\newblock Intrarow uncut weed detection using you-only-look-once instance
  segmentation for orchard plantations.
\newblock \emph{Sensors} 24, 893
\bibAnnoteFile{sampurno2024intrarow}

\bibitem[{Subhash(2024)}]{Javeri-dataset}
[Dataset] Subhash, S. (2024).
\newblock Javeri dataset.
\newblock \url{ https://universe.roboflow.com/sai-subhash/javeri }.
\newblock Visited on 2024-09-17
\bibAnnoteFile{Javeri-dataset}

\bibitem[{Szegedy et~al.(2017)Szegedy, Ioffe, Vanhoucke, and
  Alemi}]{szegedy2017inception}
Szegedy, C., Ioffe, S., Vanhoucke, V., and Alemi, A. (2017).
\newblock Inception-v4, inception-resnet and the impact of residual connections
  on learning.
\newblock In \emph{Proceedings of the AAAI conference on artificial
  intelligence}. vol.~31
\bibAnnoteFile{szegedy2017inception}

\bibitem[{Szegedy et~al.(2015)Szegedy, Liu, Jia, Sermanet, Reed, Anguelov
  et~al.}]{szegedy2015going}
Szegedy, C., Liu, W., Jia, Y., Sermanet, P., Reed, S., Anguelov, D., et~al.
  (2015).
\newblock Going deeper with convolutions.
\newblock In \emph{Proceedings of the IEEE conference on computer vision and
  pattern recognition}. 1--9
\bibAnnoteFile{szegedy2015going}

\bibitem[{Szegedy et~al.(2016)Szegedy, Vanhoucke, Ioffe, Shlens, and
  Wojna}]{szegedy2016rethinking}
Szegedy, C., Vanhoucke, V., Ioffe, S., Shlens, J., and Wojna, Z. (2016).
\newblock Rethinking the inception architecture for computer vision.
\newblock In \emph{Proceedings of the IEEE conference on computer vision and
  pattern recognition}. 2818--2826
\bibAnnoteFile{szegedy2016rethinking}

\bibitem[{Thakuria and Erkinbaev(2023)}]{thakuria2023improving}
Thakuria, A. and Erkinbaev, C. (2023).
\newblock Improving the network architecture of yolov7 to achieve real-time
  grading of canola based on kernel health.
\newblock \emph{Smart Agricultural Technology} 5, 100300
\bibAnnoteFile{thakuria2023improving}

\bibitem[{Wang et~al.(2024{\natexlab{a}})Wang, Chen, Liu, Chen, Lin, Han
  et~al.}]{wang2024yolov10}
Wang, A., Chen, H., Liu, L., Chen, K., Lin, Z., Han, J., et~al.
  (2024{\natexlab{a}}).
\newblock Yolov10: Real-time end-to-end object detection.
\newblock \emph{arXiv preprint arXiv:2405.14458}
\bibAnnoteFile{wang2024yolov10}

\bibitem[{Wang et~al.(2023{\natexlab{a}})Wang, Li, Han, Wu, and
  Zou}]{wang2023performance}
Wang, C., Li, C., Han, Q., Wu, F., and Zou, X. (2023{\natexlab{a}}).
\newblock A performance analysis of a litchi picking robot system for actively
  removing obstructions, using an artificial intelligence algorithm.
\newblock \emph{Agronomy} 13, 2795
\bibAnnoteFile{wang2023performance}

\bibitem[{Wang et~al.(2024{\natexlab{b}})Wang, Yeh, and Liao}]{wang2024yolov9}
Wang, C.-Y., Yeh, I.-H., and Liao, H.-Y.~M. (2024{\natexlab{b}}).
\newblock Yolov9: Learning what you want to learn using programmable gradient
  information.
\newblock \emph{arXiv preprint arXiv:2402.13616}
\bibAnnoteFile{wang2024yolov9}

\bibitem[{Wang et~al.(2020{\natexlab{a}})Wang, Wu, Zhu, Li, Zuo, and
  Hu}]{wang2020eca}
Wang, Q., Wu, B., Zhu, P., Li, P., Zuo, W., and Hu, Q. (2020{\natexlab{a}}).
\newblock Eca-net: Efficient channel attention for deep convolutional neural
  networks.
\newblock In \emph{Proceedings of the IEEE/CVF conference on computer vision
  and pattern recognition}. 11534--11542
\bibAnnoteFile{wang2020eca}

\bibitem[{Wang et~al.(2023{\natexlab{b}})Wang, Bao, Dong, Bjorck, Peng, Liu
  et~al.}]{wang2023image}
Wang, W., Bao, H., Dong, L., Bjorck, J., Peng, Z., Liu, Q., et~al.
  (2023{\natexlab{b}}).
\newblock Image as a foreign language: Beit pretraining for vision and
  vision-language tasks.
\newblock In \emph{Proceedings of the IEEE/CVF Conference on Computer Vision
  and Pattern Recognition}. 19175--19186
\bibAnnoteFile{wang2023image}

\bibitem[{Wang et~al.(2020{\natexlab{b}})Wang, Kong, Shen, Jiang, and
  Li}]{wang2020solo}
Wang, X., Kong, T., Shen, C., Jiang, Y., and Li, L. (2020{\natexlab{b}}).
\newblock Solo: Segmenting objects by locations.
\newblock In \emph{Computer Vision--ECCV 2020: 16th European Conference,
  Glasgow, UK, August 23--28, 2020, Proceedings, Part XVIII 16} (Springer),
  649--665
\bibAnnoteFile{wang2020solo}

\bibitem[{Woo et~al.(2018)Woo, Park, Lee, and Kweon}]{woo2018cbam}
Woo, S., Park, J., Lee, J.-Y., and Kweon, I.~S. (2018).
\newblock Cbam: Convolutional block attention module.
\newblock In \emph{Proceedings of the European conference on computer vision
  (ECCV)}. 3--19
\bibAnnoteFile{woo2018cbam}

\bibitem[{Xie et~al.(2021)Xie, Wang, Yu, Anandkumar, Alvarez, and
  Luo}]{xie2021segformer}
Xie, E., Wang, W., Yu, Z., Anandkumar, A., Alvarez, J.~M., and Luo, P. (2021).
\newblock Segformer: Simple and efficient design for semantic segmentation with
  transformers.
\newblock \emph{Advances in neural information processing systems} 34,
  12077--12090
\bibAnnoteFile{xie2021segformer}

\bibitem[{Yang et~al.(2022)Yang, Zhu, Liu, and Wang}]{yang2022land}
Yang, Q., Zhu, Y., Liu, L., and Wang, F. (2022).
\newblock Land tenure stability and adoption intensity of sustainable
  agricultural practices in banana production in china.
\newblock \emph{Journal of Cleaner Production} 338, 130553
\bibAnnoteFile{yang2022land}

\bibitem[{Yue et~al.(2023)Yue, Qi, Na, Zhang, Liu, and Liu}]{yue2023improved}
Yue, X., Qi, K., Na, X., Zhang, Y., Liu, Y., and Liu, C. (2023).
\newblock Improved yolov8-seg network for instance segmentation of healthy and
  diseased tomato plants in the growth stage.
\newblock \emph{Agriculture} 13, 1643
\bibAnnoteFile{yue2023improved}

\bibitem[{Zhang et~al.(2023)Zhang, Han, Qiao, Kim, Bae, Lee
  et~al.}]{zhang2023faster}
Zhang, C., Han, D., Qiao, Y., Kim, J.~U., Bae, S.-H., Lee, S., et~al. (2023).
\newblock Faster segment anything: Towards lightweight sam for mobile
  applications.
\newblock \emph{arXiv preprint arXiv:2306.14289}
\bibAnnoteFile{zhang2023faster}

\bibitem[{Zhang et~al.(2022)Zhang, Li, Ba, Lyu, Zhang, and
  Li}]{zhang2022banana}
Zhang, S., Li, X., Ba, Y., Lyu, X., Zhang, M., and Li, M. (2022).
\newblock Banana fusarium wilt disease detection by supervised and unsupervised
  methods from uav-based multispectral imagery.
\newblock \emph{Remote Sensing} 14, 1231
\bibAnnoteFile{zhang2022banana}

\bibitem[{Zhang et~al.(2021)Zhang, Pang, Chen, and Loy}]{zhang2021k}
Zhang, W., Pang, J., Chen, K., and Loy, C.~C. (2021).
\newblock K-net: Towards unified image segmentation.
\newblock \emph{Advances in Neural Information Processing Systems} 34,
  10326--10338
\bibAnnoteFile{zhang2021k}

\bibitem[{Zhao et~al.(2017)Zhao, Shi, Qi, Wang, and Jia}]{zhao2017pyramid}
Zhao, H., Shi, J., Qi, X., Wang, X., and Jia, J. (2017).
\newblock Pyramid scene parsing network.
\newblock In \emph{Proceedings of the IEEE conference on computer vision and
  pattern recognition}. 2881--2890
\bibAnnoteFile{zhao2017pyramid}

\bibitem[{Zhao et~al.(2023)Zhao, Ding, An, Du, Yu, Li et~al.}]{zhao2023fast}
Zhao, X., Ding, W., An, Y., Du, Y., Yu, T., Li, M., et~al. (2023).
\newblock Fast segment anything.
\newblock \emph{arXiv preprint arXiv:2306.12156}
\bibAnnoteFile{zhao2023fast}

\bibitem[{Zhou et~al.(2019)Zhou, Zhao, Puig, Xiao, Fidler, Barriuso
  et~al.}]{zhou2019semantic}
Zhou, B., Zhao, H., Puig, X., Xiao, T., Fidler, S., Barriuso, A., et~al.
  (2019).
\newblock Semantic understanding of scenes through the ade20k dataset.
\newblock \emph{International Journal of Computer Vision} 127, 302--321
\bibAnnoteFile{zhou2019semantic}

\end{thebibliography}

\end{document}